\documentclass[a4paper,10pt,twoside,leqno]{article}

\usepackage{clin}        
\usepackage{harvard}     

\usepackage[colorlinks=true, linkcolor=blue, citecolor=blue, urlcolor=blue, breaklinks]{hyperref}
\usepackage{enumitem}
\usepackage{relsize}
\usepackage[leqno]{amsmath}
\usepackage{amssymb}
\usepackage{xcolor}
\usepackage{booktabs}
\usepackage{colortbl}
\usepackage{caption}
\usepackage{subcaption}
\usepackage{naturaltableau}
\usepackage{tabularx}
\usepackage{multirow}
\usepackage{mathtools} 
\usepackage{proof}
\usepackage{linguex}
\usepackage{pgfplots}
\usepackage{calc}
\usepackage{hhline}
\usepackage{pgfbasepatterns}
\usepackage[e]{esvect} 
\usetikzlibrary{patterns,arrows}
\newcommand{\term}[1]{\ensuremath{\mathsf{#1}}}
\newcommand{\sns}[1]{\ensuremath{\mathsf{#1}}}
\newcommand{\lvl}[1]{\texttt{#1}}
\newcommand{\exws}{\kern3mm} 
\newcommand{\li}{\multimap}
\newcommand{\type}[1]{\ensuremath{\mathnormal{#1}}}
\newcommand{\types}{\type{\mathcal{T}}}
\newcommand{\atoms}{\type{\mathcal{A}}}
\newcommand{\np}{\type{np}}
\newcommand{\vp}{\type{vp}}
\newcommand{\adj}{\type{a\kern-1pt dj}}
\newcommand{\n}{\type{n}}
\newcommand{\pr}{\type{pr}}
\newcommand{\pp}{\type{pp}}
\newcommand{\ppart}{\type{part}}
\newcommand{\pron}{\type{pron}}
\newcommand{\s}{\type{s}}
\newcommand{\ahi}{\type{ahi}}
\newcommand{\ti}{\type{ti}}
\newcommand{\ww}{\type{ww}}

\newcommand{\x}{\type{\alpha}}
\newcommand{\red}[1]{\textcolor{red}{#1}}

\newcommand{\logo}{LangPro\,$\Sigma$\raisebox{0pt}[0pt][0pt]{${}^2$}}

\newcommand{\problem}[1]{\ensuremath{\mathtt{#1}}}
\newcommand{\extx}[1]{``{#1}''}
\newcommand{\rulevpr}{$\btimes$\,\rulen{v-pr}}
\newcommand{\rulezijn}{$\btimes\bot$}
\newcommand{\hmc}[1]{\cellcolor{black!#1}{#1}} 
\newcommand{\focus}[1]{{\smaller[1]\textcolor{black!50!red}{\textbf{#1}}}}
\newcommand{\cntxt}[1]{{\smaller[1]\textcolor{black!80}{#1}}}
\newcommand{\github}{\url{git.io/JzdGd}}

\definecolor{pal1}{HTML}{ADD9F4}    
\definecolor{pal2}{HTML}{E2C044}    
\definecolor{pal3}{HTML}{709176}    
\definecolor{pal4}{HTML}{FAB3A9}    
\definecolor{pal5}{HTML}{574D68}    

\begin{document}


\title{
    A Logic-Based Framework for \\
    Natural Language Inference in Dutch
}

\author{Lasha Abzianidze \email{l.abzianidze@uu.nl}\\
{\normalsize \bf Konstantinos Kogkalidis} \email{k.kogkalidis@uu.nl}\\
\AND \addr{UiL OTS, Utrecht University, Utrecht, the Netherlands}
}

\maketitle\thispagestyle{empty} 


\begin{abstract}

We present a framework for deriving inference relations between Dutch sentence pairs.
The proposed framework relies on logic-based reasoning to produce inspectable proofs leading up to inference labels; its judgements are therefore transparent and formally verifiable.
At its core, the system is powered by two $\lambda$-calculi, used as syntactic and semantic theories, respectively.
Sentences are first converted to \textit{syntactic} proofs and terms of the linear $\lambda$-calculus using a choice of two parsers: an Alpino-based pipeline, or Neural Proof Nets.
The syntactic terms are then converted to \textit{semantic} terms of the simply typed $\lambda$-calculus, via a set of hand designed type- and term-level transformations.
Pairs of semantic terms are then fed to an automated theorem prover for natural logic which reasons with them while using the lexical relations found in the Open Dutch WordNet.
We evaluate the reasoning pipeline on the recently created Dutch natural language inference dataset, and achieve promising results, remaining only within a $1.1$--$3.2\%$ performance margin to strong neural baselines.
To the best of our knowledge, the reasoning pipeline is the first logic-based system for Dutch.
The code is available at \github.
\end{abstract}

\section{Introduction}
\label{sec:intro}

Among the many Natural Language Understanding tasks, Natural Language Inference (NLI) is of particular interest.
An NLI task can be broadly summarised as follows: given two natural language utterances, a \textit{premise} and a \textit{hypothesis}, decide whether the former entails, contradicts, or is neutral with respect to the latter.
An NLI system requires the capacity for manipulating syntactic structure as well as lexical meaning, necessitating a holistic approach to yield meaningful results.

In recent years, the advent of neural models has set new benchmarks for NLI tasks, with general-purpose language models based on the pre-train and fine-tune paradigm claiming the lion's share in the literature.
Despite their undisputable performance, such models suffer from a variety of downsides, opaqueness and unpredictability being the most striking~\cite{sanchez-etal-2018-behavior,glockner-etal-2018-breaking,mccoy-etal-2019-right}.
In practical terms, a neural system might achieve high accuracy scores, but provides limited insight on how it arrived to a decision, thus prohibiting manual verification of the inference process and its outcome.
At the same time, the high expressiveness of modern neural architectures makes them prone to detect and capitalise on subtle statistical patterns and annotation artifacts common in popular NLI datasets~\cite{gururangan-etal-2018-annotation,poliak-etal-2018-hypothesis,tsuchiya-2018-performance}, artificially inflating their performance within the evaluation domain, but failing to generalise on out-of-distribution inputs.
These issues are further pronounced in use cases where reliability, robustness, and interpretability are of major importance.

In stark contrast to neural models, logic-based methods for NLI boast transparency, reliability and formal rigor, often at the cost of a drop in performance.
Logic-based systems offer not just a prediction, but rather the full explanation behind it, allowing a deep inspection of their inner workings that extends beyond mere quantitative comparisons.
They are also much more reliable in their non-neutral (i.e. entailment or contradiction) predictions than neural NLI systems.
Witness to this is the fact that \hypertarget{disagreements}{disagreements} between a logic-based model and a dataset's ground truth annotations are often due to noise or errors in the latter.

In this work, we utilise the SICK dataset~\cite{marelli-etal-2014-sick} in its recent Dutch translation~\cite{wijnholds2021sicknl} as an experimental test bed for the first cross-lingual application of LangPro~\cite{abzianidze-2017-langpro}, a Natural Tableau-based theorem prover.
Our inputs to the prover are semantic expressions in higher-order logic based on simple type theory.
To obtain semantic expressions, we transform typelogical grammar derivations procured from two wide-coverage parsers: a pipeline based on Alpino~\cite{alpino}, and Neural Proof Nets~\cite{npn}.
Following careful tuning, the prover employs the above expressions, in combination with the lexical semantic relations found in the Open Dutch WordNet database~\cite{ODWN:2016}, to generate logical inferences for the dataset, achieving a final accuracy of about $79\%$.

The structure of the paper is as follows.
In~\S\ref{sec:background} we initiate the unfamiliar reader to the formal systems we are utilising, namely typelogical grammars, $\lambda$-calculi and natural tableaux.
In~\S\ref{sec:methodology} we move to a more practical territory, describing the tools and processes we use and detailing each of our framework's components.
We then describe our experiments and expose our results on SICK-NL in~\S\ref{sec:experiments}, and make our assessments based on extensive qualitative error analysis in~\S\ref{sec:analysis}.
We draw comparisons to related work in~\S\ref{sec:related_work}.
The last section is reserved for some conclusive remarks and suggestions for future directions.

\section{Background}
\label{sec:background}
In this section, we provide a brief expository note to the formal systems we employ at each step of our inference pipeline.

We begin by detailing our syntactic framework of choice in~\S\ref{subsec:syntax}. 
Motivated by the need for a transparent syntax-semantics interface, we employ a semantically-geared typelogical grammar~\cite{morrill2012type,moot2012logic}. 
Typelogical grammars are rooted in the logical tradition of formal linguistics.
One of their biggest appeals is their affinity to semantic expressions due to the propositions-as-types interpretation that equates proofs with programs and propositions with functional types.
As such, they make for an ideal candidate in the application envisaged here.

We then describe the logic used for linguistic semantics in~\S\ref{subsec:semantics}.
The use of formal logic to model natural language inference is common practice in formal semantics. 
The usual suspect is first-order predicate logic, due to its well-behavedness, on the one hand, and the wide accessibility of out-of-the-box automated theorem provers and model builders, on the other~\cite{BlackburnBos:2005,Blackburn2001}.
Despite its attractiveness at a first glimpse, capturing meaning with first-order logic formulas can be notoriously difficult (common pain points include, among others, the representation of phenomena involving subsective adjectives and generalised quantifiers).
We therefore opt for a higher-order logic in the form of a simply-typed $\lambda$-calculus.
Aside from its expressive power, it boasts a clear syntax that resembles linguistic expressions; as such, its terms are easy to obtain from parse structures.

We conclude the section by describing the procedure of reasoning with natural language sentences in \S\ref{subsec:inference}.
Our inferential engine is powered by Natural Tableau~\cite{muskens:10,abzianidzethesis}, as inspired by the Natural Logic project~\cite{benthem:viewOnNatLog:2008,Moss2010}, a study of reasoning with meaning representations close to natural language, and the semantic tableau method, one of the most popular proof procedures for formal logics~\cite{Tableau:1999}.

\subsection{Syntax}
\label{subsec:syntax}

Typelogical grammars use a \textit{lexicon} to assign \textit{types} to words and model parse structures as logical \textit{proofs} built with the aid of a small set of inference rules.
Our typelogical grammar's logical backbone is the implication-only fragment of Intuitionistic Linear Logic~\cite{wadler1993taste}.
Its types $\types$ are inductively defined as a set closed under a single binary operator, i.e. they form a magma $( \atoms, \li )$, where:
\begin{itemize}
    \item $\atoms \subset \types$ a finite set of basic or \textit{atomic} types
    \item $\li$ the linear implication (or lolli), an operator such that $\tau_1, \tau_2 \in \types \Longleftrightarrow \type{\tau}_1 \li \type{\tau}_2 \in \types$
\end{itemize}
A type of the form $\tau_1 \li \tau_2$ (shorthand: $\type{\tau}_1(\type{\tau}_2)$) is called \textit{complex}, and is used to denote a linear transformation that will consume an \textit{argument} of type $\type{\tau}_1$ to produce a \textit{result} of type $\type{\tau}_2$.
In the linguistic setup, words that can stand on their own are assigned atomic types, whereas words requiring complements are assigned complex types; a simplified but representative lexicon is depicted in Table~\ref{tab:example_lexicon}.
    
\begin{table}[h]
    \centering
    \begin{tabularx}{0.65\textwidth}{@{}lccXlcc@{}}
        ganzen, eenden, bessen  & :: & $\np$          & & je, me  & :: & $\pron$ \\
        zwemmen                 & :: & $\np(\s)$      & & gaf     & :: & $\pron(\np(\pron(\s)))$\\    
        eten                    & :: & $\np(\np(\s))$ & & die     & :: & $(\pron(\s))(\np(\np))$ \\
        rode, blauwe            & :: & $\np(\np)$     & & en      & :: & $\forall \x: \x(\x(\x))$
    \end{tabularx}
    \caption{A toy lexicon of simple linear types.}
    \label{tab:example_lexicon}
\end{table}

In its simplest form, the type logic provides three rules of inference, through which complex expressions can be built from simple ones. 
Owing to the remarkable equivalence between logics and $\lambda$-calculi known as the Curry-Howard correspondence~\cite{sorensen2006lectures}, each logical rule has an analogue in the \textit{term} language of the linear $\lambda$-calculus:
\[
\infer[\li E]{\Gamma, \Delta \vdash \term{(s~t)}^{\type{\tau}_2}}{
    \Gamma \vdash \term{s}^{\type{\tau}_1(\type{\tau}_2)} 
    & \Delta \vdash \term{t}^{\type{\tau}_1}
}
\qquad
\infer[Ax]{\term{x}^\type{\tau} \vdash \term{x}^\type{\tau}}{}
\qquad
\infer[\li I]{\Gamma \vdash \left(\term{\lambda x.s}\right)^{\type{\tau}_1(\type{\tau}_2)}}{
    \Gamma, \term{x}^{\type{\tau}_1} \vdash \term{s}^{\type{\tau}_2}}
\]

The implication elimination rule ($\li E$) posits that, given the derivability  of an expression $\term{s}$ of type $\type{\tau}_1(\type{\tau}_2)$ from some context $\Gamma$ and the derivability of an expression $\term{t}$ of type $\type{\tau}_1$ from some context $\Delta$, from the two contexts together we can derive the term $\term{s~t}$ of type $\type{\tau}_2$, corresponding to the application of $\term{s}$ to $\term{t}$.
Together with our toy lexicon, this rule already suffices to derive terms for a few simple sentences:
\ex.
    \label{ex:applications}
    \a. eenden zwemmen\\
    {
        $\term{\left({zwemmen}^{\np(\s)}~{eenden}^{\np}\right)^{\s}}$
    }
    \b. eenden eten rode bessen\\
    {
        $\term{\left({eten}^{\np(\np(\s))}~
            \left(
                {rode}^{\np(\np)}~
                {bessen}^{\np}
            \right)^{\np}~
            {eenden}^{\np}
        \right)^{\s}}$
    }
    \c. je gaf me bessen\\
    {
        $\term{\left({gaf}^{\pron(\np(\pron(\s)))}~
                    {me}^{\pron}~
                {bessen}^{\np}~
            {je}^{\pron}
        \right)^{\s}
        }$
    }

The next two rules are a crucial component of the type logic, giving us access to hypothetical reasoning, a tool required for the derivation of higher-order syntactic phenomena.
The identity axiom ($Ax$) allows us to instantiate a fresh named \textit{variable} $\term{x}$ of some type $\type{\tau}$.
Finally, if given a context $\Gamma$ and a variable $\term{x}$ of type $\type{\tau}_1$ we can derive a term $\term{s}$ of type $\type{\tau}_2$, the implication introduction rule ($\li I$) allows us to build a function $\term{\lambda x.s}$ of type $\type{\tau}_1(\type{\tau}_2)$ from context $\Gamma$ alone.
With the addition of the above rules, we can now derive terms for more complicated sentences:

\ex.
    \label{ex:relativisation}
    bessen die je me gaf\\
    {
        $\term{\left(
                {die}^{(\np(\s))(\np(\np))}~
                \left(\lambda {x}.
                    \left(
                        {gaf}^{\pron(\np(\pron(\s)))}~
                        {me}^{\pron}~
                        {x}^{\np}~
                        {je}^{\pron}
                    \right)^{\s}
                \right)^{\np(\s)}
            ~
            {bessen}^{\np}
        \right)^{\np}}$
    }

The last item in our syntactic toolshed is a hint of type polymorphism, a telling example being the type $\forall \x:\x(\x(\x))$ assigned to coordinators, where $\x$ is a \textit{variable} ranging over types.
The above recipe gives us the means to derive conjunctions of different syntactic categories in a uniform way:

\ex.
    \label{ex:conjunction}
    \a.
    \label{ex:np_conjunction}
    ganzen en eenden\\
    {
        $\term{\left(
            {en}^{\x(\x(\x))}~
            {ganzen}^{\np}~
            {eenden}^{\np}
        \right)^{\np}}$
    }
    \b.
    \label{ex:ellipsis}
    eenden eten rode en ganzen blauwe bessen \\
    {
    $
    \begin{array}{ll}
        \term{
        \Big({en}^{\x(\x(\x))}} &
        \term{\left(\lambda {xy}.
                \left(
                    {x}^{\np(\np(\s))}~({rode}^{\np(\np)}~{y}^{\np})~{eenden}^{\np}
                \right)^{\s}\right)^{(\np(\np(\s))(\np(\s))}}\\
        &
        \term{
                \left(\lambda {zw}.
                    \left(
                        {z}^{\np(\np(\s))}~({blauwe}^{\np(\np)}~{w}^{\np})~{ganzen}^{\np}
                    \right)^{\s}\right)^{(\np(\np(\s))(\np(\s))}}\\
        &
        \term{
        {eten}^{\np(\np(\s))}~{bessen}^{\np} \Big)^{\s}
        }
    \end{array}
    $
    }
    
where in Example~\ref{ex:np_conjunction} we set $\x:=np$ for a simple noun phrase conjunction, whereas in \ref{ex:ellipsis} we set $\x:=(\np(\np(\s))(\np(\s))$ for a conjunction of sentences sharing their phrasal head, but also the head of the object noun phrase.
This last example is indicative of the treatment of elliptical constructions in a linear regime.

\subsection{Semantics}
\label{subsec:semantics}

The semantic logic we opt for is a higher-order simple predicate logic corresponding to the implication-only fragment of Intuitionistic Logic, or, in Curry-Howard terms, the simply typed $\lambda$-calculus.
It is a close replica of our syntactic logic, modulo implication no longer being linear: complex types are now the type signatures of ordinary functions.
In practical terms, the type forming operator is now an arrow $\to$ rather than the lollipop $\li$, and a single rule of inference is added to our vocabulary:
\[
    \infer[Contraction]{\Gamma, \sysm{x}^{\type{\tau}_1} \vdash \term{s}[\sysm{x}/\sysm{x}_1, \sysm{x}/\sysm{x}_2]^{\type{\tau}_2}}{
        \Gamma, \sysm{x}_1^{\type{\tau}_1}, \sysm{x}_2^{\type{\tau}_1} \vdash \sysm{s}^{\type{\tau}_2}
    }
\]
It suggests that if from some context $\Gamma$ together with two distinct variables $\sysm{x}_1$, $\sysm{x}_2$ of the same type $\type{\tau}_1$ we can derive a term $\sysm{s}$ of type $\type{\tau}_2$, then we can do the same with just a single instance of $\term{x}$, provided we replace all occurrences of $\sysm{x}_1$ and $\sysm{x}_2$ in $\sysm{s}$ with $\sysm{x}$. 
The effect of a non-linear semantic logic is that our semantic terms may now contain more than a single occurrence of terms appearing just once in the corresponding syntactic term, essentially permitting duplication of words when necessary (the utility of this will become evident in Example~\ref{ex:semantic_crd} later on).

Just like syntactic terms, semantic terms (otherwise called Lambda Logical Forms, or LLFs) are built up from variables and constant lexical items.
LLFs are typed using a small set of atoms: $\np$, $\n$, $\pp$, $\pr$, and $\s_x$, corresponding to noun phrase, common noun, prepositional phrase, particle, and sentence respectively.\footnote{
    In addition to the predicative adjectives feature ($\adj$), sentential clause category can be subcategorised as declarative ($dcl$), active past participial ($pt$), passive past participial ($pss$), present participial ($ng$), or a question (e.g. $q$ or $wh$). These sentence category features are inspired by the Combinatory Categorial Grammar (CCG) treebank~\cite{ccgbank:07}. 
    \hypertarget{very_happy_license}{Note that} the category feature $\adj$ in CCG serves to prevent certain ungrammatical declarative sentence, e.g. \extx{John happy} and \extx{John very love a woman}.
}
A few sample LLFs are depicted in Example~\ref{llfs}; to distinguish between syntactic and semantic expressions, we format lexical terms of the latter as boldface.
For types we use the abbreviation $\vp := \np(\s)$.

\bgroup
\setlength{\Exlabelsep}{.3em}
\setlength{\SubExleftmargin}{1.6em}
\ex. \label{llfs}
    \a. A woman who loves John is very happy \hypertarget{very_happy}{~}\\
        $\sysm{a}^{\n(\vp(\s))} 
          \left( \sysm{who}^{\vp(\n(\n))} 
            \left( \sysm{love}^{(\np(\vp))}~ \sysm{john}^{\np} \right) \sysm{woman}^{\n}
          \right) 
          \left( \sysm{be}^{\vp_\adj(\vp_{dcl})} \left( \sysm{very}^{\vp_\adj(\vp_\adj)} \sysm{happy}^{\vp_{\adj}} \right)
          \right)$\kern-5mm
          \vspace{2mm}
    \b. \label{llfs:b}
        Every man loves a woman\\
        $\sysm{every}^{\n(\vp(\s))}~ \sysm{man}^\n ~
          \left(\abst{x} \sysm{a}^{\n(\vp(\s))}~ \sysm{woman}^\n \left(\sysm{love}^{\np(\vp)}~ x^\np\right)
          \right)$ \hfill {\smaller Object narrow scope}\\
        $\sysm{a}^{\n(\vp(\s))}~ \sysm{woman}^\n~
          \left(\abst{y}\sysm{every}^{\n(\vp(\s))}~ \sysm{man}^\n~ \left(\abst{x}\sysm{love}^{\np(\vp)}~ x^\np~ y^\np\right)
          \right)$ \hfill {\smaller Object wide scope}

\egroup

\subsection{Reasoning with Natural Tableau}
\label{subsec:inference}

\begin{figure}[t]
\centering
\tabRuleFrame{\clSubs\vphantom{$|$}}{
\begin{forest}
for tree={align=center, parent anchor=south, child anchor=north, l sep=3mm, s sep=15mm}
[{$A : [\bar{c}] : \T$}\\{$B : [\bar{c}] : \F$}\toppad{4mm}
    [{$\btimes$\vspace{-8mm}}]
]
\end{forest}}{
$A$ infers $B$}
~
\tabRuleFrame{\abstPull\toppad{3mm}}{\begin{forest}
for tree={align=center, parent anchor=south, child anchor=north, l sep=3mm, s sep=6mm}
[{$\abst{x} A: [b,\bar{c}] : \X$}
    [{$A[x\!\coloneqq b] : [\bar{c}] : \X$}]
]
\end{forest}}{}
~
\tabRuleFrame{\someF}{\begin{forest}
for tree={align=center, parent anchor=south, child anchor=north, l sep=3mm, s sep=6mm}
[{$Q ~A~ B: \elist : \F$}
    [{$A : [c] : \F$}]
    [{$B : [c] : \F$\toppad{4mm}}]
]
\end{forest}}{$c$ is existing, $Q \in \{\sysm{a, some,\ldots}\}$}
~
\tabRuleFrame{\someT}{\begin{forest}
for tree={align=center, parent anchor=south, child anchor=north, l sep=3mm, s sep=15mm}
[{$Q ~A~ B: \elist : \T$}
    [{$A : [c] : \T$\\{$B : [c] : \T$}\toppad{4mm}}]
]
\end{forest}}{\tabuc{$Q \in \{\sysm{a, some,\ldots}\}$\\$c$ is a fresh constant}}
~
\tabRuleFrame{\argPush\toppad{3mm}}{\begin{forest}
for tree={align=center, parent anchor=south, child anchor=north, l sep=3mm, s sep=6mm}
[{$A ~b: [\bar{c}] : \X$}
    [{$A: [b,\bar{c}] : \X$}]
]
\end{forest}}{}
~
\tabRuleFrame{\rulen{aux}\toppad{3mm}}{\begin{forest}
for tree={align=center, parent anchor=south, child anchor=north, l sep=3mm, s sep=6mm}
[{$A B: [\bar{c}] : \X$}
    [{$B: [\bar{c}] : \X$}]
]
\end{forest}}{$A$ is auxiliary}
~
\tabRuleFrame{\rulen{adj$^\subset_\T$}}{\begin{forest}
for tree={align=center, parent anchor=south, child anchor=north, l sep=3mm, s sep=6mm}
[{$A B: [\bar{c}] : \T$}
    [{$B: [\bar{c}] : \T$}]
]
\end{forest}}{$A$ is subsective}
~
\tabRuleFrame{\rulen{pss}\toppad{3mm}}{\begin{forest}
for tree={align=center, parent anchor=south, child anchor=north, l sep=3mm, s sep=6mm}
[{$\sysm{by}^{\np(vp_{pss})(\vp_{dcl})} b^\np\, V~ c^\np: \elist : \X$}
    [{$V^{\np(\vp)}: [c,b] : \X$}]
]
\end{forest}}{$V$ with a new type is introduced}
\caption{The (tableau) inference rules that are employed in the tableau proof of Figure~\ref{fig:intro_proof}.
Each inference rules has its name and optional constraints that are explicitly stated below the rule. $\bar{c}$ denotes a (possibly empty) list of terms.
$\X$ is a variable over the truth signs $\T$ and $\F$.
}
\label{fig:rules}
\end{figure}
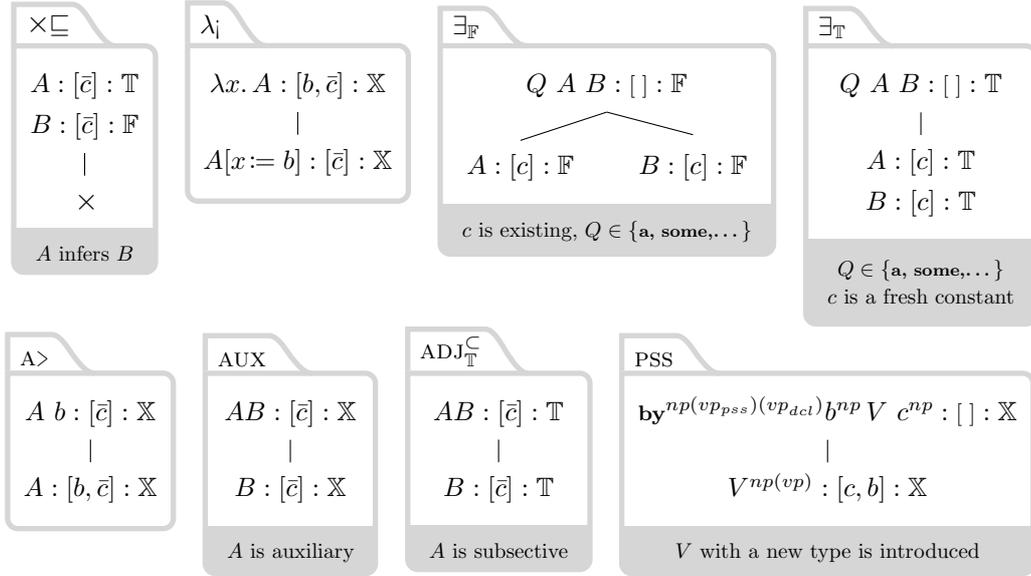

Natural Tableau is a signed tableau method specifically designed for a version of Natural Logic.
Its core component is a set of inference rules, called tableau rules.
During the reasoning process, these rules gradually break down the input logical forms, and different facets of the meaning are fleshed out.
To avoid overloaded LLFs, we will omit types of the lexical terms when appearing in tableau proofs.
Figure~\ref{fig:rules} shows some of the tableau rules.
Each rule has an antecedent and consequent entries, where a tableau entry is a triplet of a $\lambda$-term, its (possibly empty) list of arguments, and a truth sign.
For example, $A : [\bar{c}] : \T$ means that when $A$ is applied to its arguments (respecting the argument order), the resulting term is rendered as true.
Consequent entries of a rule are usually shorter than the antecedent ones, which decomposes initial terms into smaller pieces.
One special type of rule is a closure rule, e.g. (\clSubs).
The rule spots inconsistencies (like an entity being $A$ and not $B$, but at the same time $A$ being more specific than $B$) and triggers the termination of search.
The best way to understand the rules and see how they work in tandem is to consider an actual tableau proof.

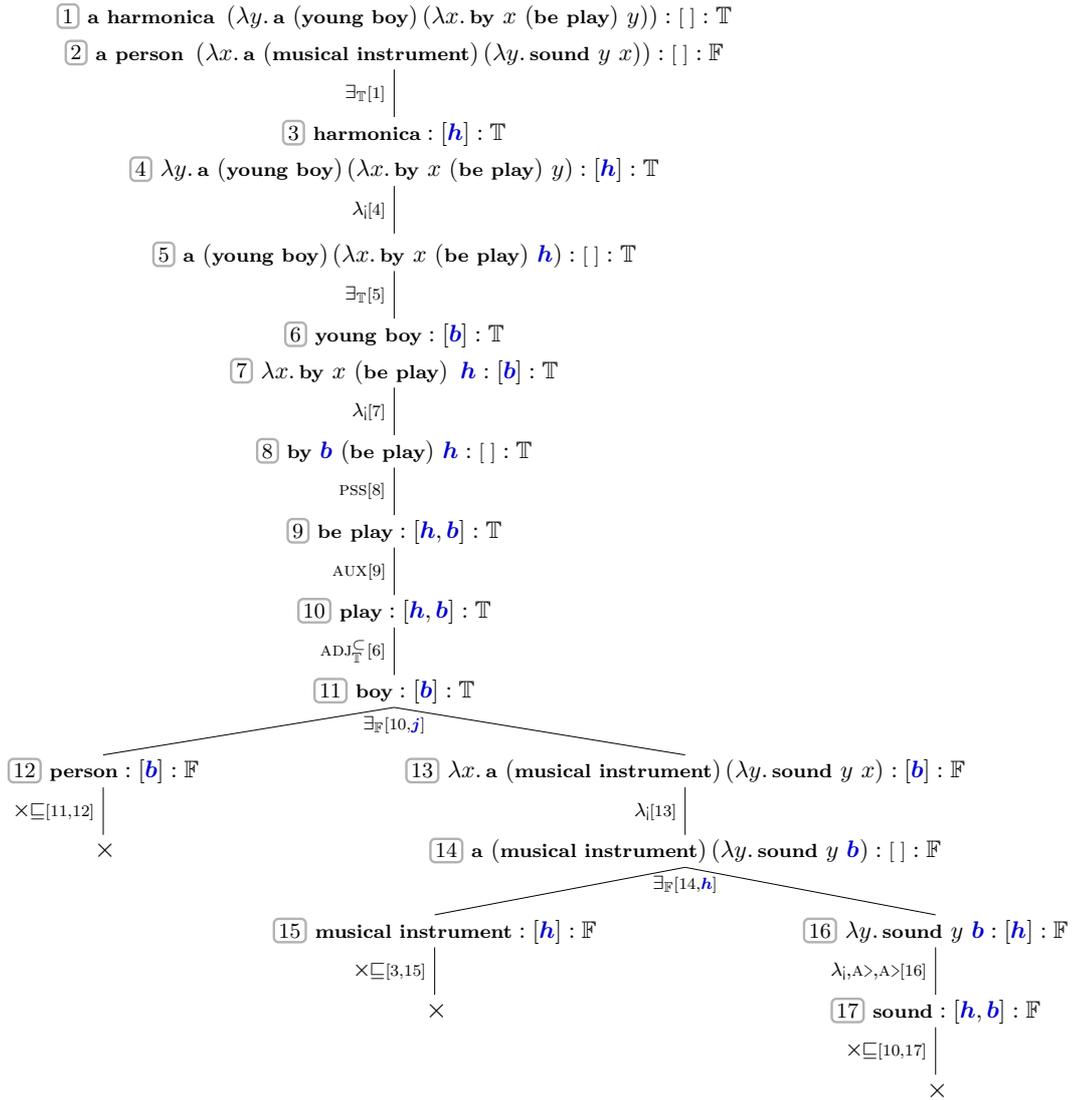
\begin{figure}[t]
\centering
\scalebox{.9}{\begin{forest}
baseline,for tree={align=center, parent anchor=south, child anchor=north, inner sep=0pt, l sep=7mm, s sep=30mm}
[{\lab{1}~$
          \sysm{a} ~ \sysm{harmonica} ~
          \left(\abst{y}\sysm{a}\,
            \left(\sysm{young} ~ \sysm{boy}
            \right)
            \left(\abst{x} \sysm{by} ~ x\, 
              \left(\sysm{be} ~ \sysm{play}
              \right)\, y
            \right)
          \right):\elist: \T$}\\
 {\lab{2}~$
    \sysm{a}~\sysm{person}~
    \left(\abst{x} \sysm{a}\,
      \left(\sysm{musical} ~ \sysm{instrument}
      \right)
      \left(\abst{y} \sysm{sound} ~ y ~ x
      \right)
    \right) :\elist: \F$}\toppad{4mm}
 [{\lab{3}~$\sysm{harmonica} : [\entity{h}] : \T$}\\
  {\lab{4}~$
    \abst{y} \sysm{a}\,
            \left(\sysm{young} ~ \sysm{boy}
            \right)
            \left(\abst{x} \sysm{by} ~ x\, 
              \left(\sysm{be} ~ \sysm{play}
              \right)\, y
            \right) : [\entity{h}] : \T$}\toppad{4mm},
  	labelA={\someT\ndList{1}},
  [{\lab{5}~$
    \sysm{a}\,
            \left(\sysm{young} ~ \sysm{boy}
            \right)
            \left(\abst{x}\sysm{by} ~ x\, 
              \left(\sysm{be} ~ \sysm{play}
              \right)\, \entity{h} 
            \right) : \elist : \T$}\toppad{4mm},
  	labelA={\abstPull\ndList{4}},
  [{\lab{6}~$\sysm{young} ~ \sysm{boy} : [\entity{b}] : \T$}\\
   {\lab{7}~$
      \abst{x} \sysm{by} ~ x\, 
              \left(\sysm{be} ~ \sysm{play}
              \right) ~\entity{h} : [\entity{b}] : \T$}\toppad{4mm},
   labelA={\someT\ndList{5}},
   [{\lab{8}~$
        \sysm{by} ~ \entity{b}\, 
              \left(\sysm{be} ~ \sysm{play}
              \right)\, \entity{h} : \elist :  \T$},
    labelA={\abstPull\ndList{7}},
    [{\lab{9}~$
        \sysm{be} ~ \sysm{play} : [\entity{h}, \entity{b}] :\T$},
     labelA={\rulen{pss}\ndList{8}},
     [{\lab{10}~$
        \sysm{play} : [\entity{h}, \entity{b}] :\T$},
      labelA={\rulen{aux}\ndList{9}},
      [{\lab{11}~$
        \sysm{boy} : [\entity{b}] :\T$},
      labelA={\rulen{adj$^\subset_\T$}\ndList{6}}, 
      labB={0mm}{\footnotesize}{\someF\ndList{10,$\entity{j}$}}
      [{\lab{12}~$
        \sysm{person} : [\entity{b}] :\F$}
       [{$\btimes$},
         labelA={\clSubs\ndList{11,12}} 
       ]
      ]
      [{\lab{13}~$
        \abst{x}\sysm{a}\,
      \left(\sysm{musical} ~ \sysm{instrument}
      \right)
      \left(\abst{y} \sysm{sound} ~ y ~ x
      \right) : [\entity{b}] : \F$},
       [{\lab{14}~$\sysm{a}\,
          \left(\sysm{musical} ~ \sysm{instrument}
          \right)
          \left(\abst{y} \sysm{sound} ~ y ~ \entity{b}
          \right) :\elist: \F$},
          labelA={\abstPull\ndList{13}},
          labB={0mm}{\footnotesize}{\someF\ndList{14,$\entity{h}$}}
        [{\lab{15}~$\sysm{musical} ~ \sysm{instrument} : [\entity{h}] : \F$}
         [{$\btimes$},
           labelA={\clSubs\ndList{3,15}} 
         ]
        ]
        [{\lab{16}~$\abst{y} \sysm{sound} ~ y ~ \entity{b} : [\entity{h}] : \F$}
         [{\lab{17}~$\sysm{sound} : [\entity{h},\entity{b}] : \F$},
          labelA={\abstPull,\argPush,\argPush\ndList{16}} 
          [{$\btimes$},
            labelA={\clSubs\ndList{10,17}} 
          ]
         ]
        ]
       ]
      ]
      ]
     ]
    ]
   ]      
  ]      
 ] 
 ]
]
\end{forest}}
\caption{The closed tableau proves that the premise \extx{a harmonica is played by a young boy} entails the hypothesis \extx{a person sounds a musical instrument}.
}
\label{fig:intro_proof}
\end{figure}

We illustrate a tableau proof in the style of Natural Tableau in Figure~\ref{fig:intro_proof}.%
\footnote{The example is in English as Natural Tableau and its computational implementation, the LangPro theorem prover, were originally developed for reasoning with English sentences.
Also following the LLF format in LangPro, lexical terms are represented as lemmas.
Additional information about types and part-of-speech (pos) tags of the lexical terms are omitted for the sake of simplicity.  
}
The tree-style proof is built to refute that the premise \extx{a harmonica is played by a young boy} entails the hypothesis \extx{a person sounds a musical instrument}. 
If the refutation fails, this serves as a proof for the entailment relation.
The refutation is carried out by searching for a counterexample for the entailment relation, i.e. building a situation that makes the premise true and the hypothesis false.
The tableau starts with this exact requirement: the LLFs of the premise and the hypothesis are set to be true and false in the nodes \lab{1} and \lab{2}, respectively.  
The rest of the tableau is built by decomposing the semantics of the initial entries with the help of the inference rules.
For example, $(\someT)$ applies to \lab{1} and produces \lab{3} and \lab{4}.
The latter produces \lab{5} via $(\abstPull)$, and so forth.
In the end, \lab{1} is decomposed as there are $b$ \extx{boy} (\lab{11}) and $h$ \extx{harmonica} (\lab{3}), and $b$ \extx{plays} $h$ (\lab{10}).
This information is inconsistent with the semantics of \lab{2}, and it is expressed in terms of three closed branches closed due to the following inconsistencies:
$b$ not being \extx{person} (\lab{12}), $h$ not being \extx{musical instrument} (\lab{15}), and $b$ not \extx{sounding} $h$ (\lab{17}).  
A tableau with its all branches closed means that the refutation failed, i.e. it was impossible to find a counterexample for the entailment relation.
Therefore, the premise entails the hypothesis; Q.E.D.

The Natural Tableau method can also be used to learn from data via abductive reasoning -- inference to the best explanation.
For example, let's assume that \extx{musical instrument} is replaced with \extx{French harp} in the hypothesis of the example in Figure~\ref{fig:intro_proof}.
The proof for entailment would fail in case there is no knowledge available saying that a \extx{harmonica} is a \extx{French harp} (e.g. such knowledge is not available in WordNet).
The idea behind abductive learning is that, given the correct/gold relation of such a problem (e.g. entailment), a tableau that attempts to prove the relation is constructed.
If the proof is not found (i.e. the tableau is not closed), then the search starts for such knowledge that helps to close the tableau.
In other words, the abductive reasoning is used to infer the knowledge that supports the correct relation.
In our example, such inferred knowledge would be \extx{harmonica} being a sort of \extx{French harp} ($\sysm{harmonica}\subs\sysm{french harp}$), which suffices to find the proof for the entailment relation.
This way, abductive learning helps infer new knowledge from labeled NLI problems, which can later be used for unseen problems.

\section{Methodology}
\label{sec:methodology}
Having introduced the formal background, we now move on to describing our implementation of the automated theorem prover for Dutch and the computational machinery behind its components.

\subsection{Parsing}
\label{subsec:parsing}

We obtain syntactic $\lambda$-terms in the form described in~\S\ref{subsec:syntax} using two different parser pipelines: one based on Alpino, and another on Neural Proof Nets.%
\footnote{
    Both pipelines generate proofs and terms enhanced with unary type- and term-level operators that specify dependency information on top of function-argument structures; \hypertarget{forgetful}{we discard} the dependency information while the structure of proofs and terms remains the same.
    We leave utilisation of the dependency information in the context of semantic reasoning as an open question for future work.
}

\paragraph{Alpino} is an attested wide-coverage parser for unrestricted written Dutch, creating parse structures in the form of dependency graphs~\cite{alpino,van-noord-2006-last}.
Its grammar, in line with head-driven phrase structure grammars~\cite{pollard1994head}, consists of a set of manually specified phrase formation rules and a rich lexicon providing subcategorisation frames and dependency information.
The format employed (graphs, rather than trees) provides the means to capture reentrancy; nodes correspond to words and phrases (labeled with syntactic category tags), and outgoing edges denote a dependency frame containing strictly one phrasal head, and zero or more complements and adjuncts.
Due to the ambiguity inherent in frame assignment and rule application, Alpino may produce a multitude of dependency graphs for a single input sentence; these are evaluated and scored on the basis of a log-linear disambiguation model aided by a few hand-designed penalisation rules.
We convert Alpino's graphs to proofs of the syntactic type logic using the type extraction algorithm of~\citeasnoun{aethel}; the algorithm traverses the parse graph, translating syntactic categories to atomic types and iteratively casting heads (resp. adjuncts) as linear functions that consume their complements (resp. phrasal parents) for each phrase.

\paragraph{Neural Proof Nets} (NPN) is a formalism-specific neurosymbolic parser composed of three parts \cite{npn}.
First, a pretrained BERT model reads the tokenised input text and builds contextualised vectorial representations for each token~\cite{delobelle2020robbert}.
An autoregressive transformer stack then translates the encoded representations into a sequence of types aligned with each input work, handling lexical type assignment and disambiguation in context~\cite{tagger}.
Finally, a permutation module based on Sinkhorn networks~\cite{mena2018learning} uses the format of multiplicative linear logic proof nets~\cite{girard1987linear} to tackle rule applications in parallel, transforming the type sequence into a proof proper.

\subsection{Obtaining Lambda Logical Forms}
\label{subsec:parsing2llf}
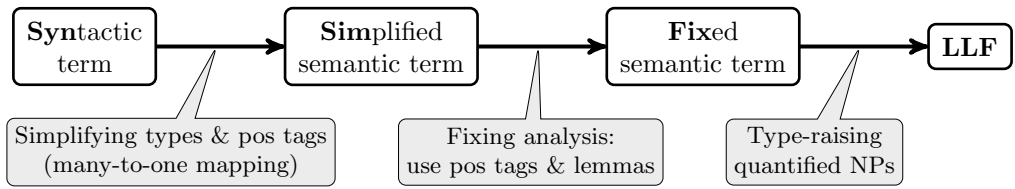
\begin{figure}[h!]
\centering
\begin{tikzpicture}
\def\bgcol{gray!15}
\def\linecol{black!70}
\tikzset{every path/.style={line width=1.5pt}}
\tikzset{every node/.style={line width=1pt, inner sep=5pt, outer sep=0pt, align=center, draw, rounded corners=1mm}}

\node(synterm)at(0,0){\textbf{Syn}tactic\\term};   

\node(simsemterm)at($(synterm.east) + (1.7,0)$)[anchor=west]
{\textbf{Sim}plified\\semantic term};  

\node(fixsemterm)at($(simsemterm.east) + (1.7,0)$)[anchor=west]
{\textbf{Fix}ed\\semantic term}; 

\node(llf)at($(fixsemterm.east) + (1.7,0)$)[anchor=west]
{\textbf{LLF}}; 

\node(simtype)at($(synterm.east)+(.2,-1.4)$)[ line width=.3pt,font=\small, inner sep=3pt, fill=\bgcol, rectangle callout, callout pointer width=2.5mm, callout absolute pointer={($(synterm.east)+(.8,0)$)}]{Simplifying types \& pos tags\\(many-to-one mapping)}; 

\node(fixing)at($(simsemterm.east)+(.7,-1.4)$)[line width=.3pt, font=\small, inner sep=3pt, fill=\bgcol, rectangle callout, callout pointer width=2mm, callout absolute pointer={($(simsemterm.east)+(.8,0)$)}]{Fixing analysis:\\use pos tags \& lemmas}; 

\node(typeraise)at($(fixsemterm.east)+(.2,-1.4)$)[line width=.3pt, font=\small, inner sep=3pt, fill=\bgcol, rectangle callout, callout pointer width=3mm, callout absolute pointer={($(fixsemterm.east)+(.8,0)$)}]{Type-raising\\quantified NPs}; 

\path [->,>=stealth'] (synterm.east) edge (simsemterm.west);

\path [->,>=stealth'] (simsemterm.east) edge (fixsemterm.west);

\path [->,>=stealth'] (fixsemterm.east) edge (llf.west);
\end{tikzpicture} 
\caption{A procedure of obtaining LLFs from the syntactic terms of the parsers.}
\label{fig:chart_llf_gen}
\end{figure}

Despite also being $\lambda$-expressions, LLFs differ from syntactic terms not only in their intended use, but also their structure and their types.
Whereas syntactic terms capture the tectogrammatical structure underlying the sentence in a bottom-up fashion, LLFs are used to express the sentential meaning, and are processed top-down by the Natural Tableau inference rules.
The conversion from the former to the latter is handled by a manually-designed pipeline, generally following~\cite{abzianidze-2015-towards}. 
The pipeline is depicted in Figure~\ref{fig:chart_llf_gen}; it gradually simplifies syntactic expressions, homogenising parser inconsistencies and syntactic subtleties that would otherwise be attenuated in the output LLFs.

\paragraph{Simplifying lexical entries}
The syntactic terms come with fine-grained types built up from 31 atoms.
On the other hand, Natural Tableau operates on LLFs typed using the following atoms: $\{\n, \np, \s_x, \pp, \pr\}$ as detailed in~\S\ref{subsec:inference}.
We translate syntactic terms into simple semantic terms using a many-to-one map from syntactic \textit{atoms} to simplified \textit{types}, as depicted in Table~\ref{tab:type_mapping}.
This serves two functions: it first collates the (quite large) set of syntactic primitives to a more manageable size, but also casts syntactic atoms that hide their semantic frames into explicit functions.

\begin{table}[h!]
    \centering
    \begin{tabularx}{0.75\textwidth}{@{} r l l @{}}
    \multicolumn{2}{c}{\textbf{Syntactic Atoms} (description \& sign)} & \textbf{LLF Types}\\
    \toprule
    Declarative sentence (verb at the 2$^\text{nd}$ position)
    & $\s_{main}$ & $\s_{dcl}$\\ 
    Subordinate clause (verb final)  
    & $\s_{sub}$  & $\s_{sub}$  \\
    Pronoun 
    & $\type{vnw}$   & \np \\
    Preposition 
    & $\type{vz}$    & \pr \\
    Numeral 
    & $\type{tw}$    & $\np$ \\
    aan het-infinitive group
    & $\ahi$   & $\np (\s_{ng})$\\
    Verb
    & $\ww$    & $\np (\s_{b})$\\
    Passive/perfect participle
    & $\ppart$ & $\np (\s_{pt})$\\
    te-infinitive group 
    & $\ti$    & \multirow{2}{*}{$\np (\s_{to})$}\\
    om te-infinitive-group  
    & $\type{oti}$   &  \\
    Adjectival Phrase 
    & $\type{ap}$    & \multirow{2}{*}{$\np (\s_{adj})$}\\
    Adjective 
    & $\adj$   & \\
    \end{tabularx}
    \caption{Mapping from typelogical atoms to LLF types.}
    \label{tab:type_mapping}
\end{table}

Syntactic constants for lexical entries are translated to semantic constants that respect the translation of their type signature.
Example~\ref{ex:type_casting} shows the simple semantic term resulting from casting each of the syntactically flat atoms $\ti$, $\ppart$ and $\ww$ to the function type $\vp := \np(\s)$.%
    \footnote{
    The simplification step also involves mapping the Alpino-style and Universal~\cite{petrov-etal-2012-universal} pos \hypertarget{spacy_foot}{tags} (coming from Alpino and spaCy, respectively, see~\S\ref{subsec:setup}) to the Penn Treebank-style.
    The latter is the tagset expected by an existing reasoning component (see \S\ref{subsec:reasoning}).  
}
\paragraph{Fixing analyses}
Adjustment and correction of simple semantic terms is the most elaborate part of the conversion procedure.
Since there is no clear cut between adjustments and corrections due to the differences in two styles of analysis motivated from top-down and bottom-up approaches, we will not distinguish them and call it fixing.
During term fixing, pos tags of the lexical entries are used in addition to type information.
Lexical terms are replaced with their lemmas (formatted in boldface). 
\hypertarget{fixing_rules}{Below} we illustrate representative instances of the term fixes for Dutch syntactic terms.%
\footnote{
Some of these fixes are results of combinations of the already existing fixing rules from \citeasnoun{abzianidze-2015-towards} specific to CCG derivation trees and the fixing rules specific to the structures of the Dutch syntactic terms.    
}

Moving a determiner in a term structure above noun modifiers is the most applied rule.
The instance of the rule application is shown in Example~\ref{ex:det_mod_precedence}.
Other fixing rules related to NPs are the rules that change the type of a verb or a preposition term that takes an argument of type $\n$.
Example~\ref{ex:n_arg} shows the changes in the types of \extx{zijn} and \extx{snijden} and the insertion of explicit quantifiers for existence and plurality for the bare NPs \extx{hout} and \extx{mannen} respectively.

Sometimes adjectives that act like nouns are analysed as predicative adjectives of type $\np (\s_{adj})$, like in (\lvl{sim}) of~\ref{ex:ss2pp_adj2n}.
The type of such nominal adjectives are set to $\n$ and the type of related lexical terms are  changed accordingly. 
Example~\ref{ex:ss2pp_adj2n} also shows how the predicative PP \extx{op een berg} is changed from an adjunct phrase to a complement of the copula \extx{zijn}.

\bgroup
\setlength{\Exlabelsep}{.5em}
\setlength{\SubExleftmargin}{1.5em}
\ex.
    \label{ex:conversions}
    \a. 
    \label{ex:type_casting}
        \a.[(\lvl{nld})\exws] om te vissen gebruikt
        \b.[(\lvl{syn})\exws] 
        {${\term{om}^{\ti(\ppart(\ppart))}~
                \left(\term{te}^{\ww(\ti)}~\term{vissen}^{\ww}\right)~
                \term{gebruikt}^{\ppart}
          }$
        }
        \c.[(\lvl{sim})\exws]
        {${\term{om}^{\vp(\vp(\vp))}~
                \left(\term{te}^{\vp(\vp)}~\term{vissen}^{\vp}\right)~
                \term{gebruikt}^{\vp}
          }$
        }
        \d.[(\lvl{fix})\exws] 
        {${\sysm{om}^{\vp(\vp(\vp))}~
                \left(\sysm{te}^{\vp(\vp)}~\sysm{vissen}^{\vp}\right)~
                \sysm{gebruiken}^{\vp}
          }$
        }\z.\vspace{3mm}
    \b. 
    \label{ex:det_mod_precedence}
        \a.[(\lvl{nld})\exws] een grote bruine hond
        \b.[(\lvl{sim})\exws]  
        {$
            \term{grote}^{\np(\np)}~
                \left(\term{bruine}^{\np(\np)}~
                    \left(
                        \term{een}^{\n(\np)}
                        ~\term{hond}^{\n}
                    \right)
                \right)
        $}
        \c.[(\lvl{fix})\exws] 
        {$
             \sysm{een}^{\n(\np)}~
                \left(\sysm{groot}^{\n(\n)}~
                    \left(
                        \sysm{bruin}^{\n(\n)}
                        ~\sysm{hond}^{\n}
                    \right)
                \right)
        $}\z.\vspace{3mm}
    \c. 
    \label{ex:n_arg}
        \a.[(\lvl{nld})\exws] mannen zijn hout aan het snijden 
        \b.[(\lvl{sim})\exws]  
        {$
            \term{zijn}^{\vp(\n(\s))}~
                \left(\term{aan\_het}^{\vp(\vp)}~
                    \left(
                        \term{snijden}^{\n(\vp)}
                        ~\term{hout}^\n
                    \right)
                \right)~
                    \term{mannen}^\n
        $}
        \c.[(\lvl{fix})\exws] 
        {$
            \sysm{zijn}^{\vp(\np(\s))}~
                \left(\sysm{aan\_het}^{\vp(\vp)}~
                    \left(
                        \sysm{snijden}^{\np(\vp)}~
                        \left(\sysm{een}^{\n(\np)} ~ \sysm{hout}^\n \right)
                    \right)
                \right)~
                \left( \sysm{s}^{\n(\np)}~ \sysm{man}^\n \right)
        $}
        \d.[(\lvl{llf})\exws]
        {$
            \sysm{s}^{\n(\vp(\s))}~\sysm{man}
            ~\left(
                \sysm{zijn}
                ~\left(\sysm{aan\_het}
                    ~\left(\abst{x}
                        \sysm{een}^{\n(\vp(\s))} ~ \sysm{hout}
                        ~\left(\abst{y}
                            \sysm{snijden}
                            ~y^\np~x^\np
                        \right)
                    \right)
                \right)
            \right)
        $}\z.\vspace{3mm}
    \d. 
    \label{ex:ss2pp_adj2n}
        \a.[(\lvl{nld})\exws] een man in het blauw is op een berg 
        \b.[(\lvl{sim})\exws]  
        {$
            \term{op}^{\np(\s(\s))}
                \left(\term{een}^{\n(\np)} ~\term{berg}^\n\right)
                \left(\term{is}^\vp
                    \left(\term{in}^{\np(\np(\np))}
                        \left(\term{het}^{\vp(\np)} ~\term{blauw}^\vp\right)
                        \left(\term{een}^{\n(\np)} ~\term{man}^\n\right)
                    \right)
                \right)
        $}
        \c.[(\lvl{fix})\exws] 
        {$
            \sysm{zijn}^{\pp(\vp)}
                \left(\sysm{op}^{\np(\pp)}
                    \left(\sysm{een}^{\n(\np)}~ \sysm{berg}^\n\right)
                \right)
                \left(\sysm{een}^{\n(\np)}~
                    \left(
                        \left(
                            \sysm{in}^{\np(\n(\n))}~
                            \left(
                                \sysm{het}^{\n(\np)}~\sysm{blauw}^\n
                            \right)
                        \right)~
                        \sysm{man}^{\n}
                    \right)
                \right)
        $}\vspace{1mm}
        \d.[(\lvl{llf})\exws] 
            \setlength\arraycolsep{0pt}
            $
            \begin{array}{ll}
                \sysm{het}^{\n(\vp(\s))} ~\sysm{blauw}~
                \big(\abst{x}
                    &\sysm{een}^{\n(\vp(\s))}~ 
                        \left(
                            \sysm{in}~x^\np ~\sysm{man}
                        \right)\\
                    &\left(\abst{y}
                            \sysm{een}^{\n(\vp(\s))}~\sysm{berg}^\n
                        \left(\abst{z}
                            \sysm{zijn}~
                            \left(\sysm{op}~z^\np\right)~ y^\np
                        \right)
                    \right)
                \big)
            \end{array}
            $
        \z.\vspace{3mm}
    \e. 
    \label{ex:semantic_crd}
        \a.[(\lvl{nld})\exws] een rode jas en kaki broek
        \b.[(\lvl{sim})\exws] 
        {$
            \term{en}^{\x(\x(\x))}~
                \term{
                \left(\abst{x}\term{rode}^{\np(\np)}      
                    \left(x^{\n(\np)}~\term{jas}^\n\right)
                \right)}~
                \term{
                \left(\abst{y}\term{kaki}^{\np(\np)}
                    \left(y^{\n(\np)}~\term{broek}^\n\right)
                \right)}~
                \term{een}^{\n(\np)}
        $}
        \c.[(\lvl{fix})\exws]  
        {$ 
            \sysm{en}^{\np(\np(\np))}
                \left(\sysm{een}^{\n(\np)}
                    \left(\sysm{rood}^{\n(\n)}~\sysm{jas}^\n\right)
                \right)
                \left(\sysm{een}^{\n(\np)}~
                    \left(\sysm{kaki}^{\n(\n)}~\sysm{broek}^\n\right)
                \right)
        $}

\egroup

Elliptical coordination constructions are modeled with syntactic terms containing $\lambda$-abstractions, as shown in (\lvl{sim}) of Example~\ref{ex:semantic_crd}.
We apply a non-linear rewriting rule to such constructions that distributes the argument over the coordinated function terms: in the example, \extx{een} is distributed over \extx{rode jas} and \extx{kaki broek}.
After the argument distribution, $\beta$-reductions are applied and the determiners are moved at the top level of NPs, as done in Example~\ref{ex:det_mod_precedence}. 

\paragraph{Type-raising NPs}
The final step in the conversion is to obtain LLFs from fixed terms.
This is done by type-raising NPs with determiners/quantifiers.
This procedure follows the algorithm described in~\citeasnoun{abzianidzethesis}, which is already implemented in the the LangPro theorem prover.
Examples of LLFs with type-raised NPs are given for the sentences~\ref{ex:n_arg} and \ref{ex:ss2pp_adj2n}.
All lexical terms retain their types except the determiners; their $\n(\np)$ type is replaced with $\n(\vp(\s))$.

\subsection{Natural Language Reasoning}
\label{subsec:reasoning}
Reasoning over the Dutch LLF is handled by LangPro~\cite{abzianidze-2017-langpro}, a Natural Tableau-based automated theorem prover.
The prover, in its original English implementation, uses a CCG parser to parse and tag input sentences, and builds two tableaux (one for entailment, and one for contradiction%
\footnote{
    To prove the contradiction relation between a premise and a hypothesis, a tableau starts with the both premise and hypothesis marked with the true sign because a counterexample for the contradiction relation is when both of the sentences can be true. 
}) 
while using the Princeton WordNet~\cite{Miller:1995} as a lexical knowledge base (KB).
LangPro has been applied to a few NLI benchmarks, and its results rank high among logic-based NLI systems~\cite{abzianidzethesis}.
In order to enable the acquisition of novel lexical knowledge from data,~\citeasnoun{abzianidze-2020-learning} recently proposed a training methodology that models learning as abductive reasoning.

We extend the theorem prover to allow processing of Dutch sentences; switching between languages can be done easily by setting the corresponding flag.
The adaptation process includes changes in two prover components:
the inventory of tableau rules, and the knowledge base.
We extend the scope of one tableau rule (\vacmod) and add a new closure rule (\rulezijn) to the rule inventory due to the analysis of Dutch expletive constructions differing from English ones.
This contrast is shown in Examples~\ref{ex:expletive_eng} and~\ref{ex:expletive_nld}.
Dutch syntactic terms treat the expletive \extx{er} as a clause modifier, while its English counterpart \extx{there} is an argument of the main verb, following the CCG analysis.
We also introduce a new rule (\rulevpr) for Dutch phrasal verbs.

\bgroup
\setlength{\Exlabelsep}{.5em}
\setlength{\SubExleftmargin}{1.5em}
\ex.
    \label{ex:expletive_eng}
    \a.[(\lvl{eng})\exws] There is no dog looking around\vspace{1mm}
    \b.[(\lvl{llf})\exws]
    {$
    \sysm{no}^{\n(\np)}~
    \left(
        \sysm{wh}^{\vp(\n(\n))}~
        \left(
            \sysm{around}^{\vp(\vp)}~\sysm{look}^\vp
        \right)~
        \sysm{dog}^\n
    \right)~
        \left(\abst{x}
        \sysm{be}^{\np(\vp)} ~ x ~ \sysm{there}^\np 
    \right)
    $}\vspace{2mm}
    
\ex.
    \label{ex:expletive_nld}
    \a.[(\lvl{nld})\exws]  Er is geen hond die rondkijkt\vspace{1mm}
    \b.[(\lvl{llf})\exws]
    {$
    \sysm{er}^{\s(\s)}~
    \left(
        \sysm{geen}^{\n(\np)} ~
        \left(
            \sysm{die}^{\vp(\n(\n))}~
            \sysm{rondkijken}^\vp~
            \sysm{hond}^\n
        \right)~
        \sysm{zijn}^\vp 
    \right)
    $}
    
\egroup 

\begin{figure}
    \centering
    \begin{minipage}[t]{70mm}
\tabRuleFrame{\vacmod\toppad{3mm}}{\begin{forest}
for tree={align=center, parent anchor=south, child anchor=north, l sep=3mm, s sep=6mm}
[{$\sysm{er}^{\s(\s)} ~ A: \elist : \X$}
    [{$A : \elist : \X$}]
]
\end{forest}}{}
\hspace{5mm}
\tabRuleFrame{\rulezijn\toppad{3mm}}{\begin{forest}
for tree={align=center, parent anchor=south, child anchor=north, l sep=3mm, s sep=6mm}
[{$\sysm{zijn}: [c] : \F$}
    [{$\btimes$}]
]
\end{forest}}{}
\\\vspace{5mm}
\\
\tabRuleFrame{\rulevpr\toppad{3mm}}{\begin{forest}
for tree={align=center, parent anchor=south, child anchor=north, l sep=3mm, s sep=6mm}
[{$A ~ R^\pr : [\bar{c}] : \X$}
    [{$R\!+\!B : [\bar{c}] : \bar{\X}$}]
]
\end{forest}}{\tabuc{if $\X = \T$, $A$ infers $B$;\\if $\X = \F$, $B$ infers $A$}}
\end{minipage}
\hspace{-30mm}
\raisebox{3mm}{
\scalebox{.9}{\begin{forest}
baseline,for tree={align=center, parent anchor=south, child anchor=north, inner sep=0pt, l sep=8mm, s sep=20mm}
[{\lab{1}~$
  \sysm{een} ~ \sysm{hond} ~ \left(\sysm{kijken} ~ \sysm{rond}\right) : \elist : \T$}\\
 {\lab{2}~$
  \sysm{er} ~
  \left(
      \sysm{geen} ~
      \left( \sysm{die} ~ \sysm{rondkijken} ~ \sysm{hond} \right)~
      \sysm{zijn} 
  \right) : \elist : \T$}\toppad{4mm}
 [{\lab{3}~$    
   \sysm{geen} ~
   \left(\sysm{die} ~ \sysm{rondkijken} ~ \sysm{hond} \right)~
   \sysm{zijn} 
   : \elist : \T$},
   labelA={\vacmod\ndList{2}},
  [{\lab{4}~$\sysm{hond} : [\entity{h}] : \T$}\\
   {\lab{5}~$\sysm{kijken} ~ \sysm{rond} : [\entity{h}] : \T$}\toppad{4mm},
    labelA={\someT\ndList{1}},
    labB={1mm}{\footnotesize}{\noT\ndList{3,$\entity{h}$}}
   [{\lab{6}~$
     \sysm{die} ~ \sysm{rondkijken} ~ \sysm{hond} : [\entity{h}] : \F$},
     labB={1mm}{\footnotesize}{\andF\ndList{6}}
    [{\lab{8}~$\sysm{rondkijken} : [\entity{h}] : \F$}
     [{$\btimes$}, labelA={\rulevpr\ndList{5,8}} 
     ]
    ]
    [{\lab{9}~$\sysm{hond} : [\entity{h}] : \F$}
     [{$\btimes$}, labelA={\clSubs\ndList{4,9}} 
     ]
    ]
   ]
   [{\lab{7}~$
     \sysm{zijn} : [\entity{h}] : \F$}
    [{$\btimes$}, labelA={\rulezijn\ndList{7}} 
    ]
   ]
  ] 
 ]
]
\end{forest}}
}
    \caption{The tableau proves the SICK-NL problem \problem{5220}: \extx{Een hond kijkt rond} contradicts \extx{Er is geen hond die rondkijkt}.
    The proof uses new rules specially designed for Dutch.
    $\bar{\X}$ is a negated version of $\X$, where $\T$'s negation is $\F$ and vice versa.}
    \label{fig:dutch_cont_proof}
\end{figure}
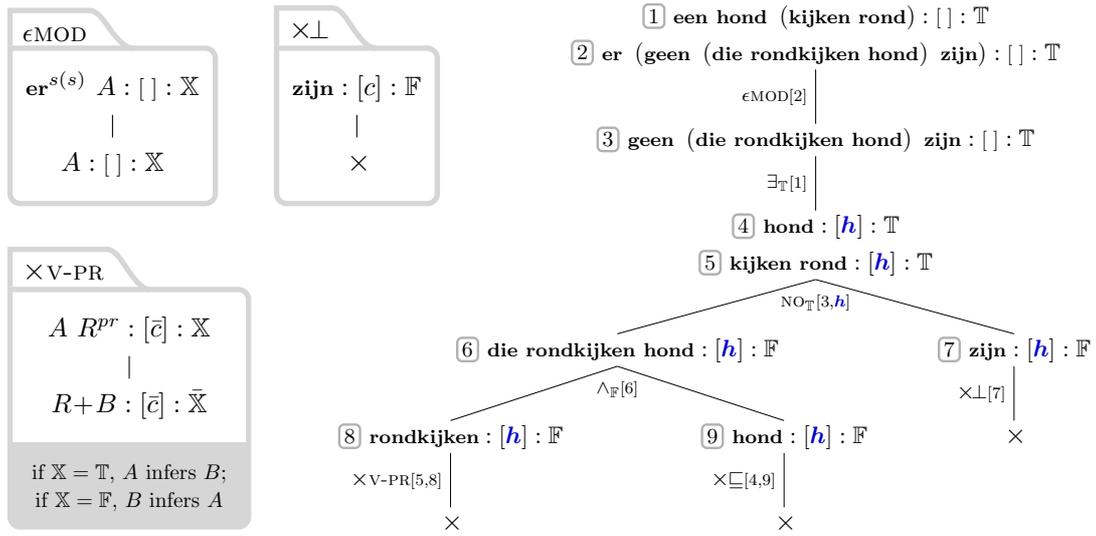

A tableau proof in Figure~\ref{fig:dutch_cont_proof} illustrates the Dutch-specific additions to the tableau rule inventory.
In order to prove that \extx{Een hond kijkt rond} contradicts \extx{Er is geen hond die rondkijkt}, the tableau method shows that there is no possible situation where both sentences are true; therefore, the tableau construction starts with entries \lab{1} and \lab{2} marked as true.
In the proof, the (\vacmod) rule treats the expletive \extx{er} as a semantically vacuous modifier (i.e. the identity function) when \lab{3} is obtained from \lab{2}.
The right-hand side branch is closed after (\rulezijn) is applied to \lab{7}.
Similarly to reasoning with English, the relative pronoun \extx{die} is analysed as a logical conjunction.
The middle branch is closed due to \lab{4} and \lab{9} contradicting each other.
The left-hand side branch is closed with the help of the new (\rulevpr) rule, which identified contradiction between \lab{5} (it is true that \extx{$h$ kijkt rond})  and \lab{8} (it is false that \extx{$h$ rondkijkt}).

Obviously, without a language-specific lexical database, the tableau prover would only be able to tackle dull logical relations.
To further allow reasoning with lexical knowledge, we employ the Open Dutch WordNet~\cite{ODWN:2016}.
To make the database \hypertarget{compatible}{compatible} with the theorem prover, we convert it to the Princeton-style prolog format.%
\footnote{
    \url{https://wordnet.princeton.edu/documentation/prologdb5wn}
}
In addition to the antonymy and hyperonymy relations present in the resource, we also use near synonymy and cross-category near synonymy, which we cast as similarity and derivational morphosemantic relation, respectively.

We adopt a default approach of the theorem prover when extracting KB relations from the WordNet.
In particular, two words are in a certain relation if there exist word senses of these words for which the corresponding WordNet relation holds.
Put differently, it is an \emph{all-sense} approach, where all corresponding word senses are considered when comparing two words.
By opting for this approach, we avoid additionally complicating our pipeline by adding a word sense disambiguation system to it.
Moreover,~\citeasnoun{abzianidze-2015-tableau} showed that it works reasonably well for the English SICK dataset.

\section{Experiments}
\label{sec:experiments}
In order to experimentally validate our methodology, we utilise SICK-NL, a recently created Dutch NLI dataset, described in~\S\ref{subsec:sick}.
We perform a range of experiments involving various combinations of tools, settings and baselines, detailed in~\S\ref{subsec:setup}, and present our results in~\S\ref{subsec:results}.

\subsection{SICK-NL}
\label{subsec:sick}
The SICK dataset~\cite{marelli-etal-2014-sick} is a collection of 6\,076 sentences originating from image captions.
These sentences are arranged in 9\,840 problems of ordered pairs, made up of a premise and a hypothesis together with an inference label (neutral, entailment or contradiction) that signifies the one-directional logical inference from premise to hypothesis.
The dataset is originally segmented in three parts: train, trial and test (4\,500, 500 and 4\,927 problems respectively) for the SemEval-14 shared task~\cite{marelli-etal-2014-semeval}.

The Dutch version of SICK~\cite{wijnholds2021sicknl} is derived from the original through an automated translation process and a gold inference label transfer.
The dataset consists of 6\,060 unique sentences in total.
The automatic translations were manually inspected, and erroneous ones were corrected on an individual sentence basis (i.e. not taking pairing contexts into account).
Based on preliminary experiments with neural models,~\citeasnoun{wijnholds2021sicknl} indicate that the Dutch counterpart is more difficult than the original; the main hypothesised reason is a reduction in lexical overlap between sentence pairs (despite a small reduction in average sentence length), owing to machine translation inflating the dataset's vocabulary.

\subsection{Experimental Setup}
\label{subsec:setup}
We experiment with syntactic expressions from both parsers.
From Alpino, we request the global optimal parse for each sentence, imposing no time constraints.
The graphs obtained are then converted into typelogical derivations with the aid of the extraction algorithm.
In the process, a small portion of parses are discarded due to either underspecifying the sentence's function/argument structure (e.g. resorting to vague discourse-level annotations) or failing rudimentary correctness checks.
From Neural Proof Nets, we select the structurally correct analysis with the highest score that falls within a beam of width 6 (when at least one such exists).
Table~\ref{tab:parser_coverage} reports sentence and dataset coverage from each parser pipeline.

\begin{table}[h]
    \centering
    \begin{tabularx}{0.7\textwidth}{@{}lXcc@{}}
        \textbf{Parser}     & & \textbf{Sentences Parsed} & \textbf{Problems Covered} \\
        \toprule
         Neural Proof Nets  & & 5\,812 \smaller{(95,9\%)} & 9\,264 \smaller{(94,1\%)} \\ 
         Alpino             & & 5\,947 \smaller{(98,1\%)} & 9\,611 \smaller{(97,7\%)}
    \end{tabularx}
    \caption{Sentences parsed and problems covered with each parser.}
    \label{tab:parser_coverage}
\end{table}

We then transform syntactic analyses to logical forms as described in~\S\ref{subsec:parsing2llf}.
In order to ease lexical lookup, we homogenise semantic constants by lemmatising them, using pos tags to disambiguate, and apply term conversions when necessary.
Lemmas and tags are obtained from two sources: Alpino, and \hypertarget{spacy}{spaCy's} large Dutch language model~\cite{spacy}.
Logical forms are then fed to the LangPro theorem prover -- if a problem is missing a term for either premise or hypothesis, the prover's prediction defaults to the neutral label.

We obtain an accuracy score on the test set (percentage of problems correctly classified) from each parser \& tagger combination, as well as five ensemble models.
Each ensemble aggregates the votes of equally weighted models, prioritising non-neutral over neutral votes, and defaulting to neutral in case of conflict (e.g. entailment vs contradiction).
We produce two ensembles over parsers, two over taggers, and one over all four taggers and parser combinations.

Next, we train each of the core model using abduction on the union of the training and trial portions of the dataset.
We use the trial set in abductive learning since the theorem prover has no proper set of hyperparameters that can be tuned in the  development phase.
For the abductive learning we use the settings of~\citeasnoun{abzianidze-2020-learning}. 
Post-training, models are organised in ensemble pairs as before, without cross-model spilling of learned knowledge.

To quantitatively assess our models' performance, we compare against established pretrained language models, fine-tuned as three-way sequence classifiers (a sequence being the concatenation of the premise and hypothesis sentences, as is standard practice).
Following~\citeasnoun{wijnholds2021sicknl}, we use BERTje~\cite{bertje}, RobBERT~\cite{delobelle2020robbert} and mBERT~\cite{bert}, but perform model selection on the basis of trial set accuracy, and average scores from five training instances.

\begin{table}[b!]
    \begin{subtable}{0.52\textwidth}
        \centering
        \begin{tabularx}{\textwidth}{
            @{}Xc@{~}cc@{~}cc@{~}c@{}}
        {} & \multicolumn{6}{c}{\textbf{Parser}} \\ 
        \addlinespace
        \textbf{Tagger} & \multicolumn{2}{c}{\textsc{npn}} & \multicolumn{2}{c}{\textsc{alpino}} & \multicolumn{2}{c}{$\Sigma$}\\
        \cmidrule{1-7}
        Alpino    &  74.65 & \smaller{-1.50} & 75.87 & \smaller{-1.83} & 76.38 & \smaller{-1.75} \\
        spaCy     &  76.66 & \smaller{-1.38} & 77.61 & \smaller{-1.72} & 78.38 & \smaller{-1.58} \\
        $\Sigma$  &  77.04 & \smaller{-1.40} & 77.98 & \smaller{-1.71} & 78.83 & \smaller{-1.62} \\
        \end{tabularx}
        \caption{Accuracy of LangPro when using each parser \& tagger combination, including ensembles ($\Sigma$) over parser, tagger or both. Right subcolumns report difference when no training with abduction is used.}
        \label{tab:xcombinations}
    \end{subtable}\hfill
    \begin{subtable}{0.42\textwidth}
        \centering
        \begin{tabularx}{\textwidth}{@{}Xcc@{}} 
             \textbf{Model} & \textbf{Accuracy} & \textbf{Hybrid} \\
             \midrule
             \logo & 78.8 & -- \\ 
             \midrule
             BERTje & 82.0 & 81.8\\
             RobBERT  & 81.7 & 82.6\\
             mBERT & 79.9 & 80.6\\
        \end{tabularx}
        \caption{Performance of \logo\ compared to fine-tuned neural baselines.
        Right column reports performance when \logo\ proofs override neural predictions.
        }
        \label{tab:baselines}
    \end{subtable}
    \caption{Internal and external model comparisons on the test set of SICK-NL, with $56.4\%$ of neutral-class baseline. The scores are percentage of correctly classified problems.}
    \label{tab:comparisons}
\end{table}

\subsection{Results}
\label{subsec:results}
Table~\ref{tab:xcombinations} presents the results for all parser \& tagger combinations and ensembles, with and without abduction.
Comparing individual components, we note that models perform better with (i) pos tags and lemmas coming from spaCy rather than Alpino, and (ii) parse structures coming from the Alpino pipeline rather than NPN.
When it comes to abduction, trained models perform consistently better across the board, raising individual model performance by $1.38$--$1.83\%$.
In line with previous work~\cite{abzianidze-2015-tableau,martinez-gomez-etal-2016-ccg2lambda}, aggregating proofs from various model combinations substantially improves results.
Our best performing model is the ensemble of four theorem provers using all cross combinations of parsers \& taggers, where each of the prover has been trained using abduction; The ensemble model achieves a raw improvement of $1.22\%$ over its best constituent (with the Alpino parser \& spaCy tagger combination).
We abbreviate this ensemble model as \logo\ and use it for subsequent comparisons.

Unsurprisingly, and as Table~\ref{tab:baselines} suggests, all BERT-based models outperform \logo, with a maximum absolute difference of $3.2\%$.
However, inspecting the confusion matrices of the systems in Table~\ref{tab:cmatrix} reveals \logo's merits, namely the high precision of its entailment and contradiction predictions.
Proofs generated by \logo\ are reliable enough to safely override most neural models' predictions, allowing the two types of systems to complement one another.
Table~\ref{tab:baselines} shows accuracy scores for all baselines, as well as hybrid models where \logo{}'s proofs (i.e. entailment and contradiction predictions) override predictions of the neural models.
\hypertarget{nn_lp_diff}{Evidently}, \logo\ can benefit RobBERT and mBERT but not BERTje.
The reason behind the latter is that according to the gold labels, \logo\ correctly proofs 39 proofs for entailment (31) and contradiction (8) problems which are wrongly classified by BERTje, but \logo\ also provides false proofs for 48 neutral problems (the false proofs are discussed in~\S\ref{ss:missing_imagined_proofs}), which outweigh the accuracy gain from the correct proofs.

The hybrid model that pairs \logo\ and RobBERT outperforms all models, surpassing the previous benchmark of BERTje.
Closer look at the predictions of \logo\ and RobBERT reveal that RobBERT benefits most from \logo{}'s proofs (93) for entailment problems compared to mBERT (71) and BERTje (31).
Table~\ref{tab:cmatrix} also shows that RobBERT is the worst among the neural baselines in predicting entailment problems, but the adoption of \logo{}'s proofs results in the best performing hybrid model.   

\begin{table}[t]
\centering
\scalebox{.9}{
    \begin{tabular}[t]{r|r|r|r|}
        \multicolumn{4}{c}{\hspace{24mm}\textbf{\logo}}\\
        \%  & E & C & N \\\hhline{~===}
        Entailment & \hmc{14.6}  & \hmc{0.1} & \hmc{14.0} \\\cline{2-4}
        Contradiction & $<$\hmc{0.1}  & \hmc{9.8} & \hmc{4.8} \\\cline{2-4}
        Neutral & \hmc{1.5}  & \hmc{0.7} & \hmc{54.5} \\\cline{2-4}
    \end{tabular}
\hspace{2mm}
    \begin{tabular}[t]{|r|r|r|}
        \multicolumn{3}{c}{\textbf{BERTje}}\\
        E & C & N \\\hline\hline
        \hmc{24.7}  & \hmc{0.1} & \hmc{3.9} \\\hline
        \hmc{0.7}  & \hmc{12.7} & \hmc{1.3} \\\hline
        \hmc{9.5}  & \hmc{2.6} & \hmc{44.6} \\\hline
    \end{tabular}\bigskip
\hspace{2mm}
    \begin{tabular}[t]{|r|r|r|}
        \multicolumn{3}{c}{\textbf{RobBERT}}\\
        E & C & N \\\hline\hline
        \hmc{22.1}  & \hmc{0.1} & \hmc{6.6} \\\hline
        \hmc{0.6}  & \hmc{12.5} & \hmc{1.5} \\\hline
        \hmc{7.2}  & \hmc{2.3} & \hmc{47.1} \\\hline
    \end{tabular}
\hspace{2mm}
    \begin{tabular}[t]{|r|r|r|}
        \multicolumn{3}{c}{\textbf{mBERT}}\\
        E & C & N \\\hline\hline
        \hmc{22.7}  & \hmc{0.2} & \hmc{5.8} \\\hline
        \hmc{0.6}  & \hmc{12.0} & \hmc{2.0} \\\hline
        \hmc{9.3}  & \hmc{2.2} & \hmc{45.2} \\\hline
    \end{tabular}
}
\caption{Confusion matrices on the test set. The numbers represent percentage of the total problems.}
\label{tab:cmatrix}
\end{table}

\begin{table}[b!]
    \centering
    \begin{tabularx}{\textwidth}{@{}rr@{~~}l@{}}
         \multicolumn{2}{l}{\textbf{id/Label}~~~} & \textbf{Sentences}\\
         \toprule
         \problem{168} & p & Een kind slaat een honkbal\\ 
         C & h & Een kind mist een honkbal \\
         \midrule
         \problem{175} & p & Een familie kijkt naar een kleine jongen die een honkbal raakt\\
         E & h & Een jongen slaat een honkbal \\
         \midrule
         \problem{1556} & p & Een man draagt een boom\\
         E & h & Een man draagt een plant \\ 
         \midrule
         \problem{897} & p & Mensen zitten op een strand vol zand bij de oceaan en genieten van een zonnige dag \\
         C & h & Er is niemand aan de wal \\ 
         \midrule
         \problem{4470} & p & Een man schopt een voetbal\\
         E & h & Een man schopt een bal \\
    \end{tabularx}
    \caption{Based on the test set, a set of problems (cherry-picked from a total of 17 problems) that were misclassified by all neural models as neutral but solved by \logo. The gold labels are abbreviated with the initial letters.}
    \label{tab:neural_bad}
\end{table}

It is interesting to see the problems that all neural models failed at, but \logo\ solved.
\hypertarget{several}{Several} of these problems are shown in Table~\ref{tab:neural_bad}.
The problems seem easy, but for some reason all the neural baselines predict them as neutral.
It is even more mysterious how all of them predict the comparable problem \problem{169} correctly, which has the same premise as \problem{175} paired with the hypothesis \extx{Een familie kijkt naar een jongen die een honkbal slaat}.
The reasoning capacity required to predict \problem{169} correctly is sufficient for solving \problem{175}; we hypothesise that neural models give more weight to word sequence similarity when it comes to predicting entailment.

\section{Analysis}
\label{sec:analysis}
To gain a better insight on the model's performance, we perform extensive qualitative analyses targeted at either specific components of the framework (\hypertarget{explicate_subsections}{namely}, the syntactic parsers in \S\ref{ss:alpino_vs_npn} and the abductive learning in \S\ref{ss:abduction}) or particular cases of interest (missing and imagined proofs by \logo{} in \S\ref{ss:missing_imagined_proofs}).
\hypertarget{train_set}{All} the conducted analyses are based on the training part of SICK-NL to avoid eyeballing the problems from the test set.

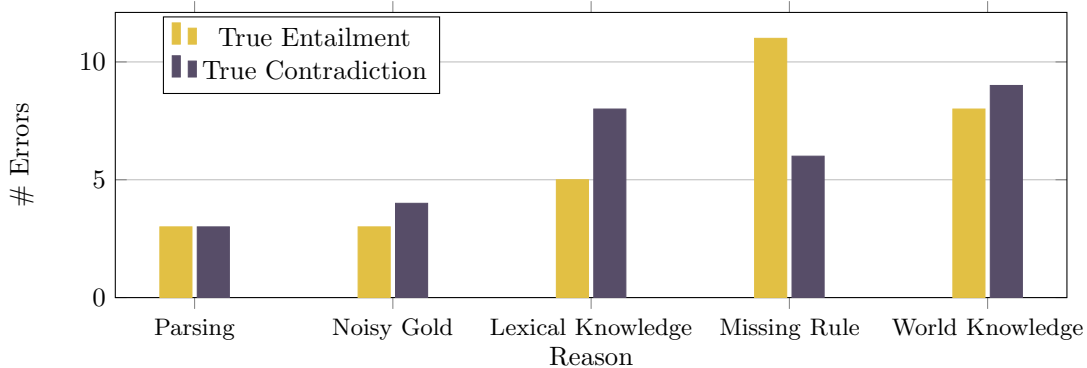
\begin{figure}[t]
    \centering
    \begin{tikzpicture}
    \begin{axis}[
        ybar,
        ymajorgrids=true,
        xlabel={Reason},
        ylabel={\# Errors},
        symbolic x coords={
            Parsing,Noisy Gold,Lexical Knowledge,Missing Rule,World Knowledge},
        xticklabel style={font=\small},
        xtick=data,
        ymin=0,
        enlarge x limits=0.1,
        width=0.925\textwidth,
        height=0.2484\textheight,
        bar width=12pt,
        legend style={at={(0.05, 0.85)}, anchor=west, legend columns=1},
    ]
    \addplot[color=pal2, fill=pal2] coordinates {
        (World Knowledge,8)
        (Missing Rule,11)
        (Lexical Knowledge,5)
        (Noisy Gold,3)
        (Parsing,3)
    }; 
    \addplot[color=pal5, fill=pal5] coordinates {
        (World Knowledge,9)
        (Missing Rule,6)
        (Lexical Knowledge,8)
        (Noisy Gold,4)
        (Parsing,3)
    }; 
    \legend{True Entailment, True Contradiction};
  \end{axis}
\end{tikzpicture}
    \caption{Error analysis for a sample of problems falsely classified as neutral.}
    \label{fig:typeII}
\end{figure}

\subsection{Missing and Imagined Proofs}\label{ss:missing_imagined_proofs}
We begin by investigating the predictions of LangPro that differ from gold labels, treating the neutral label as the null hypothesis. 
To avoid confounding our analysis with the effects of abduction, we inspect predictions by untrained LangPro versions.

\begin{table}[b!]
    \centering
    \begin{tabularx}{\textwidth}{@{}rr@{~~}l@{}}
         \multicolumn{2}{l}{\textbf{id/Label}~~~} & \textbf{Sentences}\\
         \toprule
         \problem{4701} & p & Een groep mannen \red{voetbalt} op het strand\\ 
         E & h & Een groep mannen \red{speelt met een bal} op het strand \\
         \midrule
         \problem{8073} & p & {\small Twee jongens in witte outfits en rode beschermende kleding \red{staan te sparren} op een mat}\\
         E & h & {\small Twee kinderen in witte outfits en rode beschermende kleding \red{sparren} op een mat} \\
         \midrule
         \problem{7375} & p & Een man \red{staat op de top van de rotsen} met wolken erachter\\
         N & h & Een persoon \red{zit op een bergtop} \\ 
    \end{tabularx}
    \caption{Problems that were misclassified by \logo. The gold labels are abbreviated with the initial letters.}
    \label{tab:langpro_bad}
\end{table}

\begin{figure}[t!]
    \centering
    \begin{tikzpicture}
    \begin{axis}[
        ybar,
        ymajorgrids=true,
        xlabel={Reason},
        ylabel={\# Errors},
        symbolic x coords={
            Noisy Gold,Lexical Knowledge},
        xticklabel style={font=\small},
        xtick=data,
        ymin=0,
        width=0.8\textwidth,
        enlarge x limits=0.5,
        width=0.925\textwidth,
        height=0.2484\textheight,
        bar width=26pt,
        legend style={at={(0.05, 0.85)}, anchor=west, legend columns=1},
    ]
    \addplot[color=pal2, fill=pal2] coordinates {
        (Lexical Knowledge,18)
        (Noisy Gold,3)
    }; 
    \addplot[color=pal5, fill=pal5] coordinates {
        (Lexical Knowledge,2)
        (Noisy Gold,8)
    };
    \legend{False Entailment, False Contradiction};
  \end{axis}
\end{tikzpicture}
    \caption{Error analysis for a sample of neutral problems falsely classified.}
    \label{fig:typeI}
\end{figure}
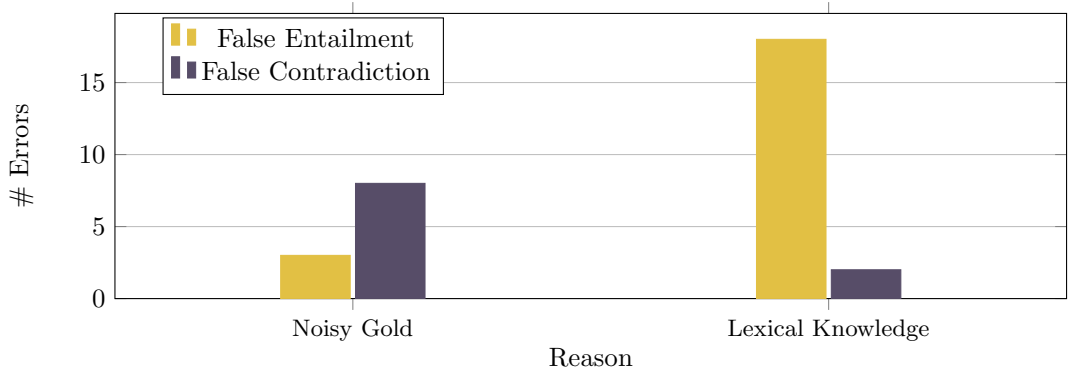

We first focus on \textit{missing proofs} or type II errors, i.e. cases where \textit{none} of the stand-alone models produce a proof (that is, a neutral prediction for a non-neutral gold label problem).
We detect a total of 1\,038 such cases out of 4\,500, and randomly sample 60 of those, \hypertarget{evenly_split}{half} between entailments and contradictions.
Our findings are presented in Figure~\ref{fig:typeII} and illustrating examples in Table~\ref{tab:langpro_bad}.
The majority of missing proofs (17) can be attributed to requiring commonsense reasoning or world knowledge, which goes beyond the capacities of the prover (e.g. problem~\problem{4701} requires knowing that \extx{voetbalt} implies \extx{speelt met een bal}).
LangPro is also responsible for not delivering equally many proofs, the reason being the absence or malfunction of a structure altering rule (e.g. \problem{8073} requires ignoring the auxiliary to equate the present continuous \extx{staan te sparren} with the simple continuous \extx{sparren}).
Another 13 cases can be explained as requiring lexical relations not present in the KB, and 7 more are due to noisy (erroneous or ambiguous) gold labels, partially caused by translation-induced meaning shifts (e.g. problem \problem{5282} of Table~\ref{tab:translation_shifts}).
Finally, 6 are to be blamed on absurd parses.

\begin{table}[!b]
    \centering
    \newcolumntype{L}{>{\raggedright\arraybackslash}X}
    \begin{tabularx}{0.9675\textwidth}{@{}l@{~~}l@{~~}lLc@{}}
        \multicolumn{3}{c}{\textbf{id}} & \multicolumn{1}{c}{\textbf{Sentences}} & \textbf{Label Change} \\
        \toprule
        \multirow{5}{*}{\rotatebox[origin=c]{90}{\problem{1202}}} &
        \multirow{2}{*}{en}  & p & A man is making a speech on a podium & 
        \multirow{2}{*}{N}\\
        & & h & A man is speaking on a stage 
        & \multirow{2}{*}{\rotatebox[origin=c]{270}{$\longmapsto$}}\\ 
        & \multirow{2}{*}{nl} & p & Een man houdt een toespraak op een podium &
        \multirow{2}{*}{?E}\\
        & & h & Een man spreekt op een podium\\
        \midrule
        \multirow{4}{*}{\rotatebox[origin=c]{90}{\problem{1818}}} &
        \multirow{2}{*}{en}  & p & Some bells are ringing near a cook slicing peppers & 
        \multirow{2}{*}{N}\\
        & & h & A cook is slicing some bell peppers 
        & \multirow{2}{*}{\rotatebox[origin=c]{270}{$\longmapsto$}}\\
        & \multirow{2}{*}{nl} & p & Er rinkelen wat belletjes bij een kok die paprika's in reepjes snijdt &
        \multirow{2}{*}{?E}\\
        & & h & Een kok snijdt wat paprika's in reepjes\\
        \midrule
        \multirow{4}{*}{\rotatebox[origin=c]{90}{\problem{5282}}} &
        \multirow{2}{*}{en}  & p & A rabbit is playing with a stuffed bunny & 
        \multirow{2}{*}{E}\\
        & & h & A bunny is playing with a stuffed bunny 
        & \multirow{2}{*}{\rotatebox[origin=c]{270}{$\longmapsto$}}\\
        & \multirow{2}{*}{nl} & p & Een konijn speelt met een knuffelkonijn &
        \multirow{2}{*}{?N}\\
        & & h & Een konijn speelt met een knuffelhaasje\\
    \end{tabularx}
    \caption{Example problems with potential label discrepancies between English and Dutch.}
    \label{tab:translation_shifts}
\end{table}

Next we turn our attention to \textit{imagined proofs} or type I errors, i.e. cases where \textit{all} of the stand-alone models produce a proof, whereas none was expected (that is, a non-neutral output with a neutral gold label).
We detect and inspect a total of 33 such cases and present our findings in Figure~\ref{fig:typeI}; most of the errors (21) are due to inaccurate lexical relations with 11 of the remaining being debatable gold labels (Table~\ref{tab:translation_shifts}).

Based on the above findings, we draw a number of conclusions pertaining to the framework as well as the dataset, and identify recurring patterns in the kinds of sentences we fail to properly analyze.
First, we emphasise the 2 orders of magnitude difference between type I \& II errors: there are almost 1\,000 cases of missing proofs, but only about 30 cases of wrong proofs, which serves to show that \logo\ is precise in its proofs but lacks high coverage, and thus makes for a good candidate first model in a hierarchical classification pipeline.

Witness to that, we remark that $33\%$ of the false positives and $12\%$ of the false negatives encountered are in fact plausible or outright correct, the issue lying with the label rather than the prediction!
A portion of the mislabeled problems found are in agreement with prior analyses of the original SICK, common causes being a lack of an absolute reference frame, no clear distinction between alteration and contradiction, ungrammatical sentences, and annotation errors~\cite{kalouli2017}.
Others, however, are unique to the Dutch translation, and can be pinpointed to the lexical choices of the machine translation system employed; Table~\ref{tab:translation_shifts} presents a few telling examples.
In several cases (problems~\problem{1202} and~\problem{1818}), two distinct source words are translated to the same target word, creating slight meaning shifts that affects the inference label.
The issue is not exclusive to type I errors; in fact, shifts occur even more frequently the other way around, translating the same word differently depending on (sometimes irrelevant) context (problem~\problem{5282}).
Albeit not always catastrophic, translation-induced inconsistencies magnify the dataset's difficulty: lexical inconsistencies increase the vocabulary size, and therefore the demand on the knowledge base, whereas grammatical inconsistencies necessitate a more exhaustive set of structure-altering rules.

Concerning Open Dutch WordNet, we note that the resource places a severe upper boundary on system performance, as 25\% of the missing proofs are due to the absence of a needed lexical relation.
This is not surprising, considering the scale of the database and the highly demanding nature of the task.
What is, however, surprising is the frequency of relations that lead to unexpected proofs.
Upon closer inspection, we distinguish two error cases.
The first is due to relations that are just plain wrong, a striking example being \sysm{kat}\subs\sysm{hond}, which contributes to a total of six imagined proofs.%
    \footnote{In the Open Dutch WordNet, while \sns{hond.n.01} is a direct hyponym of \sns{huisdier.n.01}, erroneously it is also a hypernym of it. This makes \sns{kat.n.01} a hyponym of \sns{hond.n.01} as it is a direct hyponym of \sns{huisdier.n.01}.   
    }
The other is more deeply rooted, and is associated with the all-sense approach we chose to adopt for simplicity (see~\S\ref{subsec:reasoning}).
For instance, based on the standard (i.e. frequently used) senses, \hypertarget{action}{verbs} like \textit{liggen}, \textit{lopen}, and \textit{staan} are not hyponyms of \textit{zitten}.
But there is a sense of \textit{zitten}, which means to occupy a certain position or area, and it is a hypernym of some senses of \textit{liggen}, \textit{lopen}, and \textit{staan}.
This makes the all-sense approach to adopt the relations like \sysm{liggen}\subs\sysm{zitten}, \sysm{lopen}\subs\sysm{zitten}, and \sysm{staan}\subs\sysm{zitten} and to prove problems like \problem{7375} in Table~\ref{tab:langpro_bad}.%
\footnote{It is possible to block such unwanted relations from the Open Dutch WordNet by specifically discarding certain problematic senses and hypernymy relations from it. But pushing the performance score as high as possible is not the main goal of the current paper.}



Finally, the prevalence of separable verbs in Dutch can also be a source of stress for all components of our framework.
Starting from the parsers, there is an apparent tension between optional adverbs and necessary but free-floating particles.
LangPro itself then requires careful tuning on the treatment of each case (while accounting for possible errors) before deferring to the lexical database.
The latter may often contain a relation between verbal cores but lack one for the full verbs, making derivations possible only if one selectively ignores particles; this, however, carries the danger of ignoring crucial parts of the sentential meaning.

\subsection{Alpino vs. NPN}\label{ss:alpino_vs_npn}
\begin{figure}[!t]
    \begin{tikzpicture}
    \begin{axis}[
        xbar,
        xmajorgrids=true,
        ylabel={Error Type},
        xlabel={\# Errors},
        ylabel style={at={(axis description cs:-0.05,1.05), anchor=north}, rotate=270},
        symbolic y coords={Other,Absurd Parse,Mislabeled Mod, PP Attachment, No Parse},
        yticklabel style={font=\small},
        ytick=data,
        xmin=0,
        enlarge y limits=0.175,
        width=0.9\textwidth,
        height=0.345\textheight,
        legend style={at={(0.85, 0.1)}, anchor=south, legend columns=1},
        reverse legend
    ]
    \addplot[color=pal1, fill=pal1] coordinates {
        (5,No Parse)
        (11,PP Attachment)
        (8,Mislabeled Mod)
        (4,Absurd Parse)
        (0,Other)
    }; 
    \addplot[color=pal3, fill=pal3] coordinates {
        (14,No Parse) 
        (0,PP Attachment)
        (3,Mislabeled Mod)
        (2,Absurd Parse)
        (3,Other)
    };
    \legend{Alpino,NPN}
  \end{axis}
\end{tikzpicture}
    \caption{Error analysis for a sample of problems with parser disagreements.}
    \label{fig:parser_errors}
\end{figure}
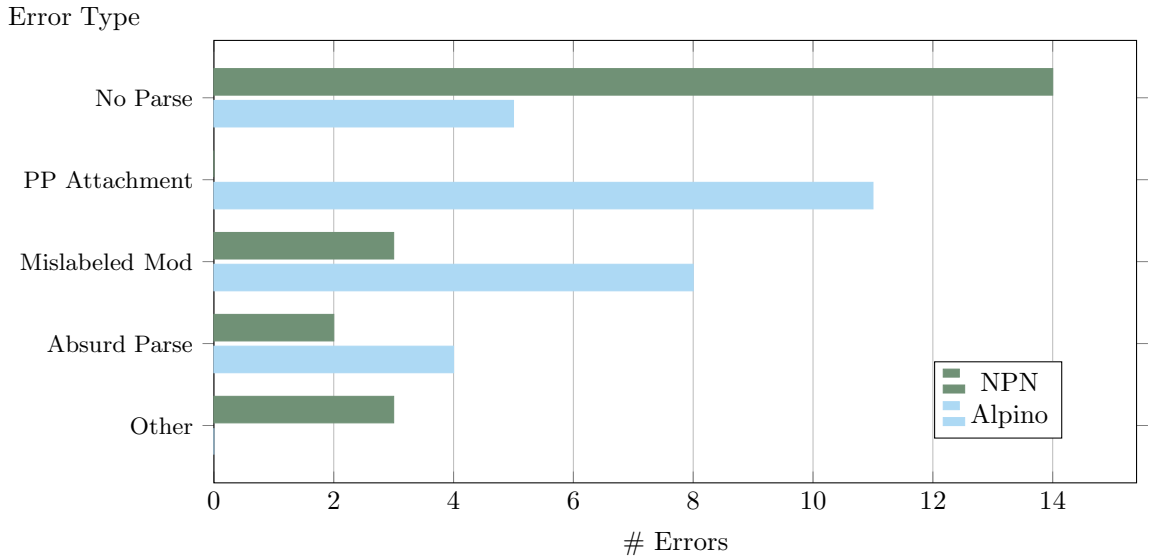

Our next analysis seeks to investigate the effect of parser choice on model performance.
Perhaps strikingly, predictions that rely on the Alpino-based pipeline seem generally more reliable than predictions based on the NPN parser, despite the latter reportedly achieving a higher parsing accuracy~\cite{npn}.
We randomly sample and inspect 50 problems where exactly one of the parsers' outputs leads to a proof.
Our findings are presented in Figure~\ref{fig:parser_errors}.
Both systems fail with almost equal frequency (26 failures from the Alpino-based pipeline and 24 from NPN), but the error sources are quite different between the two.

More than half (14) of the NPN failures arise from a lack of a parse.
Of those, 5 problems contain comma separated non-restrictive relative clauses (NPN has been trained with punctuation-free sentences), 3 contain very short and simple \extx{aan het} constructions (incorrectly analysed in NPN's training data) and 3 more contain simple transitive sentences where the subject is a conjunction (we hypothesise those to be training artifacts).
Alpino, on the other hand, rarely fails at producing any parse (5 cases), but its output is more often wrong.
There are 11 cases of PP attachment gone wrong, 8 cases of a modifier mislabeled as a predicate or vice-versa and 4 cases of severe issues in the predicted function/argument structure; NPN, in comparison, has 0, 3 and 2 of those respectively.
Considering also Table~\ref{tab:parser_coverage}, the verdict is that Alpino is more robust, boasting a higher coverage which in turn leads to more solved problems, but NPN is more accurate, boasting a higher proportion of correct parses; this serves to further justify our decision to ensemble the two, and is in line with the benefits observed when doing so.

\subsection{Abduction}\label{ss:abduction}

The abductive learning further boosts the performance of the theorem prover with the help of lexical knowledge induced during the training phase (see Table~\ref{tab:xcombinations}).
The learning bias of the abduction is to find the smallest set of relations over short phrases that explains (i.e. helps prove) the gold inference label for a premise-hypothesis pair.
Here, we manually check and analyse the relations learned by the abduction from the train and trial parts of SICK-NL.

\begin{table}[t]
    \centering
    \begin{tabularx}{\textwidth}{@{}l@{}c@{\kern-2mm}r X@{}}
        \textbf{Type} & \textbf{Count} &  (\textbf{\%}) & \multicolumn{1}{c}{\textbf{Examples of learned relations}}
        \\\midrule
        Correct & 53 & (32.9) &
        \tabul{
        \focus{zwart $\subs$ donker}: \cntxt{zwarte hond $\subs$ donkere hond}\\
        \focus{leeg $|$ vol}: \cntxt{schaatsbaan is leeg $|$ schaatsbaan is vol met mensen}\\
        \focus{liggen in het gras $|$ rennen}: \cntxt{hond ligt in het gras $|$ hond rentin het gras}\\
        \focus{pizza $\subs$ voedsel}: \cntxt{er is geen man die voedsel eet $|$ Een man eet een pizza}\\
        \focus{lopen $\subs$ rennen}: \cntxt{honden lopen snel samen $\subs$ honden rennen samen}\\
        }
        \\\midrule
        Contextual & 20 & (12.4) &
        \tabul{
        \focus{zwaard $\subs$ mes}: \cntxt{snijdt een laars met een zwaard $\subs$ snijdt een laars met een mes}\\
        \focus{boren $|$ sluiten}: \cntxt{boort een gat $|$ sluit een gat}\\
        \focus{klein $\subs$ jong}: \cntxt{kleine jongen speelt $\subs$ jonge jongen speelt}\\
        }
        \\\midrule
        Reversed & 26 & (16.1) &
        \tabul{
        \focus{container $\subs$ plastic container}: \cntxt{een container van plastic $\subs$ een plastic container}\\
        \focus{hond $\subs$ bulldog}: \cntxt{aap borstelt de hond $|$ aap borstelt geen bulldog}\\
        \focus{persoon $\subs$ fietser}: \cntxt{Iemand die op fietsen rijdt $\subs$ een fietser}\\
        }
        \\\midrule
       Prepositional & 27 & (16.8) & 
        \tabul{
        \focus{op $\subs$ door}: \cntxt{stel loopt het gangpad op $\subs$ stel loopt door het gangpad}\\
        \focus{in $\subs$ aan}: \cntxt{puppy knaagt aan een houten paal $\subs$ puppy bijt in een paal}\\
        \focus{buiten $\subs$ rond}: \cntxt{tijger loopt buiten een kooi $\subs$ tijger loopt rond een kooi}\\
        }        
        \\\midrule
        Wrong & 35 & (21.7) &
        \tabul{
        \focus{wissen $\subs$ weten}: \cntxt{man is het werk \ldots aan het wissen $\subs$ man wist het werk \ldots}\\
        \focus{suiker $\subs$ kruid}: \cntxt{voegt suiker toe aan het vlees $\subs$ voegt kruiden toe aan wat vlees}\\
        \focus{spannen $\subs$ draaien}: \cntxt{meisje \ldots spant een lint $|$ geen meisje \ldots dat een lintje draait}\\
        \focus{gieten $|$ halen}: \cntxt{giet olie in een koekenpan $|$ haalt de olie uit een koekenpan}\\
        }            
        \\\midrule
    \end{tabularx}
    \caption{Types of relations learned with the abduction and their counts. Each relation instance comes with a context from the original problem as seen during abductive learning.}
    \label{tab:learned_relations}
\end{table}

We consider the overlap between the relations learned by each version of the theorem prover differing in terms of the parser-tagger combinations.
There are in total 161 such common relations learned.
We classify each relation according to five categories.
The \emph{correct} and \emph{wrong} categories are self-explanatory.
\emph{contextual} relations require a substantial amount of context to be considered justifiable. 
\emph{Reversed} relations are reversed versions of correct subsumption relations, while \emph{prepositional} are, as the name suggests, relations over prepositions. 
A distribution of the learned relations over the five categories with accompanying examples are shown in Table~\ref{tab:learned_relations}. 

The majority of learned relations are correct, and either resemble WordNet-like entries, such as the antonym $\sysm{leeg} \disj \sysm{vol}$ and the hyper/hyponym $\sysm{pizza} \disj \sysm{voedsel}$, or commonsense-like relations, e.g. $\sysm{zwart} \subs \sysm{donker}$. 
The wrong relations cover examples caused by wrong lemmatisation (e.g. $\sysm{wissen} \subs \sysm{weten}$), noisy gold labels (e.g. $\sysm{suiker} \subs \sysm{kruid}$), and failure to correctly identify verb particle constrictions  (e.g. $\sysm{gieten} \disj \sysm{halen}$).
Worth discussing are also reversed relations.
While some relations (e.g. $\sysm{hond} \subs \sysm{bulldog}$) are induced from noisy gold labels, most of them are due to the learning bias of the abduction preferring relations with short phrases.
This preference opts for learning $\sysm{persoon} \subs \sysm{fietser}$ rather than $\sysm{persoon die op fietsen rijdt} \subs \sysm{fietser}$.

\section{Related Work}\label{sec:related_work}

There are a few logic-based systems that have been successfully applied to NLI. 
Good entry datasets for logic-based systems to NLI represent SICK \cite{marelli-etal-2014-sick} and FraCaS \cite{fracas} as the former contains relatively simple sentences mainly requiring reasoning with lexical and compositional knowledge, while the latter covers multi- and single-premised problems presupposing complex logic-based reasoning.
Two logic-based NLI systems that stand out with their performance on these NLI datasets are ccg2lambda \cite{martinez-gomez-etal-2016-ccg2lambda,yanaka-etal-2018-acquisition} and LangPro \cite{abzianidze-2017-langpro,abzianidze-2020-learning}.
While both systems are logic-based and use CCG parsers as a starting point, they differ in terms of the logical representations of sentences and the reasoning procedures.
While ccg2lambda has already been applied to Japanese NLI \cite{mineshima-etal-2016-building}, the current work represents the first cross-lingual application of LangPro.
We believe that ccg2lambda could also be adapted to Dutch likewise LangPro to Japanese as both systems require comparable resources of syntactic parsing and lexical knowledge.

Categorial grammar-based syntactic trees represent a smooth starting point when it comes to obtaining logical forms of sentences since the grammars' transparent syntax-semantic interface facilitates the meaning composition.  
That's why most of the logic-based NLI systems have used CCG-based parsers, which are the best performing categorial grammar-based parsers for English.
In this paper, we take advantage of a wide-coverage typelogical parser for Dutch, and shift our syntactic representations accordingly~\cite{kogkalidis-etal-2020-neural}.

An NLI dataset has been recently made available for Dutch \cite{wijnholds2021sicknl}, and during the writing this paper we are not aware of any Dutch NLI systems (other than the baseline neural models discussed in our experiments). 
We would like to stress here that a Dutch system for semantic similarity~\cite{marsi-krahmer-2010-automatic} is not comparable to Dutch NLI systems as it is tackling a different NLP task.
Additionally, it is measuring the similarity of syntactic trees without really reasoning about the meaning
(the latter can also be attributed to the Dutch neural NLI models to some extent).%
\footnote{ 
We include this comparison after one of the reviewers considered our work similar to the DAESO project in terns of a scope and an aim.
}
Obviously one could adapt a semantic similarity-based system to NLI, but such an endeavour would offer little in terms of explainability compared to our reasoning-based framework. 




\section{Conclusion}
Building on existing work, we have proposed a framework for the logical analysis of textual inference in Dutch.
Our work has been motivated by the recent release of a Dutch translation of the SICK dataset~\cite{wijnholds2021sicknl}, and is the first work specifically targeted to the dataset.
As our entry point, we used the two available tools for acquiring type-theoretic analysis of written Dutch to parse the entirety of the dataset~\cite{aethel,npn}.
Relying on the clean syntax-semantics interface offered by typelogical grammars and their close affinity to $\lambda$-calculi, we hand-designed a conversion scheme that first simplifies syntactic terms, before then casting them to semantic expressions.
We then employed a high-horsepower Natural Tableau prover~\cite{abzianidze-2017-langpro}, and expanded upon it with new rules, aimed at addressing some of the quirks of Dutch.
Supported with the lexical relations from the Open Dutch WordNet~\cite{ODWN:2016}, the prover first learns domain-specific relations from the training data via an abductive learning component~\cite{abzianidze-2020-learning} and then predicts unseen problems based on formal proofs.
We finally compared our system to strong neural baselines~\cite{bert,bertje,delobelle2020robbert}, and find them to be not only on an equal standing, but also complementary to one another to some extent.

Contrary to neural alternatives, our proposed framework constitutes a ``glass box'' model, yielding answers not in the form of plain labels, but rather proofs which can be both human-inspected and machine-verified. 
This allows us to gain deeper insights pertaining to the problem in all of its aspects, including the peculiarities of the dataset (e.g. detecting noisy gold labels), the strengths and weaknesses of the components employed (e.g. identifying missing relations in the Open Dutch WordNet), and the methodological decisions we have followed.
We have conducted an in-depth error analysis, which has shed light on the effects of automatic translation on the difficulty of the dataset, of word variation on lexical database stress, of parse choice on prominent error cases, and of the abduction on the quality of lexical knowledge extracted.

The extensive error analysis showed several directions for future work that can further improve our system's performance.
First, to surmount the issues related to wrong parses or inconsistent PP attachments, one could additionally consider logical forms generated from n-best parses from NPN and Alpino and/or employ a yet another syntactic parser for Dutch (e.g. a Dutch CCG parser developed as part of the \hypertarget{pmb}{Parallel Meaning Bank} \cite{abzianidze-etal-2017-parallel} is an obvious candidate for this).
Second, to improve the abductive learning, a promising direction would be to incorporate embedding-based word similarity during training, e.g. give priority to relations with a higher cosine similarity between their arguments.
Third, to better process Dutch particle verbs during term conversion and theorem proving, one could include specialised processing of verb lemmas that include a particle.
Fourth, to procure more training data~\cite{yanaka-etal-2019-help} or phenomenon-specific evaluation sets~\cite{yanaka-etal-2019-neural,RichardsonHMS20,yanaka-etal-2021-exploring} for Dutch NLI, one could automatically create labeled NLI pairs where our logic-based NLI system could serve as an integral component of the problem generation or as a sanity checker of the inference labels.  
Error analysis aside, we are finally curious to explore how dependency relations (provided by both parsing frameworks, but not utilised here) can find use to term processing, lexical disambiguation and reasoning as a whole.

In hopes that our work will prove useful to future research on Dutch NLI and acknowledging the contributions of others that made it possible, we open source our code and make it available at \github.


\section*{Acknowledgements}
We thank the organisers of the Natural Logic Meets Machine Learning (NaLoMA) workshop for hosting us, the reviewers for their suggestions on earlier drafts of this work, and the participants for attending.
We also thank the members of the Utrecht NLP Reading Group for providing a friendly environment for discussions and rehearsals.
Lasha is supported by the European Research Council (ERC) under the European Unions Horizon 2020 research and innovation programme (grant agreement No. 742204).
Konstantinos is supported by the Dutch Research Council (NWO) through the project “A composition calculus for vector-based semantic modelling with a localisation for Dutch” (360-89-070).

\bibliography{bibliography}
\bibliographystyle{clin}

\end{document}